\documentclass[10pt,twocolumn,letterpaper]{article}

\usepackage{cvpr}              
\usepackage{multirow} 

\usepackage[dvipsnames]{xcolor}

\definecolor{cvprblue}{rgb}{0.21,0.49,0.74}
\usepackage[pagebackref,breaklinks,colorlinks,citecolor=cvprblue]{hyperref}

\usepackage{adjustbox}
\usepackage{CJK}
\usepackage[noend]{algpseudocode}
\usepackage{algorithmicx,algorithm}
\usepackage [marginal] {footmisc}
\usepackage[misc]{ifsym}
\usepackage[accsupp]{axessibility}
\counterwithin{section}{part}
\renewcommand{\thesection}{\arabic{section}}

\usepackage{amsmath,amsfonts,bm, bbm}
\usepackage{xparse}

\DeclareDocumentCommand\W{ g g }{%
        \IfNoValueTF {#1} {\mathbf{W}} {
            \IfNoValueTF {#2} {\mathbf{W}^{(#1)}}{\mathbf{W}^{(#1)}_{#2}}
        }
}

\DeclareDocumentCommand\bias{ g g }{%
        \IfNoValueTF {#1} {\mathbf{b}} {
            \IfNoValueTF {#2} {\mathbf{b}^{(#1)}}{\mathbf{b}^{(#1)}_{#2}}
        }
}

\DeclareDocumentCommand\betavar{ g g }{%
        \IfNoValueTF {#1} {\bm{\beta}} {
            \IfNoValueTF {#2} {{\bm{\beta}^{(#1)}}{}}{\bm{\beta}^{(#1)}_{#2}}
        }
}

\DeclareDocumentCommand\xivar{ g g }{%
        \IfNoValueTF {#1} {\bm{\xi}} {
            \IfNoValueTF {#2} {{\bm{\xi}^{(#1)}}{}}{\bm{\xi}^{(#1)}_{#2}}
        }
}

\DeclareDocumentCommand\xivarn{ g g }{%
        \IfNoValueTF {#1} {\bm{\xi^-}} {
            \IfNoValueTF {#2} {\bm{\xi^-}^{+(#1)}}{\bm{\xi^-}^{+(#1)}_{#2}}
        }
}

\DeclareDocumentCommand\xivarp{ g g }{%
        \IfNoValueTF {#1} {\bm{\xi^+}} {
            \IfNoValueTF {#2} {\bm{\xi^+}^{+(#1)}}{\bm{\xi^+}^{+(#1)}_{#2}}
        }
}

\DeclareDocumentCommand\nuvar{ g g }{%
        \IfNoValueTF {#1} {\bm{\nu}} {
            \IfNoValueTF {#2} {{\bm{\nu}^{(#1)}}{}}{\bm{\nu}^{(#1)}_{#2}{}}
        }
}

\DeclareDocumentCommand\hnuvar{ g g }{%
        \IfNoValueTF {#1} {\bm{\hat{\nu}}} {
            \IfNoValueTF {#2} {{\bm{\hat{\nu}}^{(#1)}}{}}{\bm{\hat{\nu}}^{(#1)}_{#2}{}}
        }
}

\DeclareDocumentCommand\muvar{ g g }{%
        \IfNoValueTF {#1} {\bm{\mu}} {
            \IfNoValueTF {#2} {{\bm{\mu}^{(#1)}}{}}{\bm{\mu}^{(#1)}_{#2}}
        }
}

\DeclareDocumentCommand\gammavar{ g g }{%
        \IfNoValueTF {#1} {\bm{\gamma}} {
            \IfNoValueTF {#2} {{\bm{\gamma}^{(#1)}}{}}{\bm{\gamma}^{(#1)}_{#2}}
        }
}

\DeclareDocumentCommand\lambdavar{ g g }{%
        \IfNoValueTF {#1} {\bm{\lambda}} {
            \IfNoValueTF {#2} {{\bm{\lambda}^{(#1)}}{}}{\bm{\lambda}^{(#1)}_{#2}}
        }
}

\DeclareDocumentCommand\tbetavar{ g g }{%
        \IfNoValueTF {#1} {{\bm{\tilde{\beta}}}} {
            \IfNoValueTF {#2} {{{\bm{\tilde{\beta}}}^{(#1)}}{}}{{{\bm{\tilde{\beta}}}^{(#1)}_{#2}}}
        }
}

\DeclareDocumentCommand\alphavar{ g g }{%
        \IfNoValueTF {#1} {\bm{\alpha}} {
            \IfNoValueTF {#2} {{\bm{\alpha}^{(#1)}}}{\bm{\alpha}^{(#1)}_{#2}}
        }
}

\DeclareDocumentCommand\D{ g g }{%
        \IfNoValueTF {#1} {\mathbf{D}} {
            \IfNoValueTF {#2} {\mathbf{D}^{(#1)}}{\mathbf{D}^{(#1)}_{#2}}
        }
}

\DeclareDocumentCommand\A{ g g }{%
        \IfNoValueTF {#1} {\mathbf{A}} {
            \IfNoValueTF {#2} {\mathbf{A}^{(#1)}}{\mathbf{A}^{(#1)}_{#2}}
        }
}

\DeclareDocumentCommand\AA{ g g }{
        \IfNoValueTF {#1} {\mathbf{\Omega}} {
            \IfNoValueTF {#2} {\mathbf{\Omega}(#1, #1)}{\mathbf{\Omega}(#1, #2)}
        }
}

\DeclareDocumentCommand\S{ g g }{%
        \IfNoValueTF {#1} {\mathbf{S}} {
            \IfNoValueTF {#2} {\mathbf{S}^{(#1)}}{\mathbf{S}^{(#1)}_{#2}}
        }
}

\DeclareDocumentCommand\K{ g g }{%
        \IfNoValueTF {#1} {\mathbf{K}} {
            \IfNoValueTF {#2} {\mathbf{K}^{(#1)}}{\mathbf{K}^{(#1)}_{#2}}
        }
}

\DeclareDocumentCommand\B{ g g }{%
        \IfNoValueTF {#1} {\mathbf{B}} {
            \IfNoValueTF {#2} {\mathbf{B}^{(#1)}}{\mathbf{B}^{(#1)}_{#2}}
        }
}

\DeclareDocumentCommand\lowerb{ g g }{%
        \IfNoValueTF {#1} {{\mathbf{\underline{b}}}} {
            \IfNoValueTF {#2} {{\mathbf{\underline{b}}}^{(#1)}}{{\mathbf{\underline{b}}}^{(#1)}_{#2}}
        }
}

\DeclareDocumentCommand\z{ g g }{%
        \IfNoValueTF {#1} {z} {
            \IfNoValueTF {#2} {z^{(#1)}}{z^{(#1)}_{#2}}
        }
}

\DeclareDocumentCommand\s{ g g }{%
        \IfNoValueTF {#1} {s} {
            \IfNoValueTF {#2} {s^{(#1)}}{s^{(#1)}_{#2}}
        }
}

\DeclareDocumentCommand\dom{ g g }{%
        \IfNoValueTF {#1} {\mathcal{S}} {
            \IfNoValueTF {#2} {\mathcal{S}_{#1}}{\mathcal{S}^{#1}_{#2}}
        }
}

\DeclareDocumentCommand\domlb{ g g }{%
        \IfNoValueTF {#1} {\mathsf{LB}} {
            \IfNoValueTF {#2} {\mathsf{LB}(\mathcal{#1})}{\mathsf{LB}(\mathcal{#1}_{#2})}
        }
}

\DeclareDocumentCommand\domub{ g g }{%
        \IfNoValueTF {#1} {\mathsf{UB}} {
            \IfNoValueTF {#2} {\mathsf{UB}(\mathcal{#1})}{\mathsf{UB}(\mathcal{#1}_{#2})}
        }
}

\DeclareDocumentCommand\uns{ g g }{%
        \IfNoValueTF {#1} {\tilde{s}} {
            \IfNoValueTF {#2} {\tilde{s}_{#1}}{s^{(#1)}_{#2}}
        }
}

\DeclareDocumentCommand\ub{ g g }{%
        \IfNoValueTF {#1} {u} {
            \IfNoValueTF {#2} {u^{(#1)}}{u^{(#1)}_{#2}}
        }
}

\DeclareDocumentCommand\lb{ g g }{%
        \IfNoValueTF {#1} {l} {
            \IfNoValueTF {#2} {l^{(#1)}}{l^{(#1)}_{#2}}
        }
}

\DeclareDocumentCommand\hz{ g g }{%
        \IfNoValueTF {#1} {\hat{z}} {
            \IfNoValueTF {#2} {\hat{z}^{(#1)}}{\hat{z}^{(#1)}_{#2}}
        }
}

\DeclareDocumentCommand\bu{ g g }{%
        \IfNoValueTF {#1} {\mathbf{u}} {
            \IfNoValueTF {#2} {\mathbf{u}^{(#1)}}{\mathbf{u}^{(#1)}_{#2}}
        }
}

\DeclareDocumentCommand\bl{ g g }{%
        \IfNoValueTF {#1} {\mathbf{l}} {
            \IfNoValueTF {#2} {\mathbf{l}^{(#1)}}{\mathbf{l}^{(#1)}_{#2}}
        }
}

\DeclareDocumentCommand\aaa{ g }{%
        \IfNoValueTF {#1} {\bm{a}} {
            {\bm{a}^{({#1})}}
        }
}

\DeclareDocumentCommand\haaa{ g }{%
        \IfNoValueTF {#1} {\bm{\hat{a}}} {
            {\bm{\hat{a}}^{({#1})}}
        }
}

\DeclareDocumentCommand\bbb{ g g }{%
        \IfNoValueTF {#1} {\mathbf{P}} {
            \IfNoValueTF {#2} {{\mathbf{P}_{#1}}}{{\mathbf{P}_{#1}^{({#2})}}}
        }
}

\DeclareDocumentCommand\hbbb{ g g }{%
        \IfNoValueTF {#1} {\mathbf{\hat{P}}} {
            \IfNoValueTF {#2} {{\mathbf{\hat{P}}_{#1}}}{{\mathbf{\hat{P}}_{#1}^{({#2})}}}
        }
}

\DeclareDocumentCommand\ccc{ g g }{%
        \IfNoValueTF {#1} {\mathbf{q}} {
            \IfNoValueTF {#2} {{\mathbf{q}_{#1}}}{{\mathbf{q}_{#1}^{(#2)}}{}}
        }
}

\DeclareDocumentCommand\constc{ g }{%
        \IfNoValueTF {#1} {c} {
            {c^{({#1})}}
        }
}

\DeclareDocumentCommand\setz{ g g }{%
        \IfNoValueTF {#1} {\mathcal{Z}} {
            \IfNoValueTF {#2} {\mathcal{Z}^{(#1)}}{\mathcal{Z}^{(#1)}_{#2}}
        }
}

\DeclareDocumentCommand\setzp{ g g }{%
        \IfNoValueTF {#1} {\mathcal{Z^+}} {
            \IfNoValueTF {#2} {\mathcal{Z}^{+(#1)}}{\mathcal{Z}^{+(#1)}_{#2}}
        }
}

\DeclareDocumentCommand\setzn{ g g }{%
        \IfNoValueTF {#1} {\mathcal{Z^-}} {
            \IfNoValueTF {#2} {\mathcal{Z}^{-(#1)}}{\mathcal{Z}^{-(#1)}_{#2}}
        }
}

\DeclareDocumentCommand\tsetz{ g g }{%
        \IfNoValueTF {#1} {\tilde{\mathcal{Z}}} {
            \IfNoValueTF {#2} {\tilde{\mathcal{Z}}^{(#1)}}{\tilde{\mathcal{Z}}^{(#1)}_{#2}}
        }
}

\DeclareDocumentCommand\tz{ g g }{%
        \IfNoValueTF {#1} {\tilde{z}} {
            \IfNoValueTF {#2} {\tilde{z}^{(#1)}}{\tilde{z}^{(#1)}_{#2}}
        }
}

\DeclareDocumentCommand\f{ g g }{%
        \IfNoValueTF {#1} {f} {
            \IfNoValueTF {#2} {f^{(#1)}}{f^{(#1)}_{#2}}
        }
}

\DeclareDocumentCommand\lf{ g g }{%
        \IfNoValueTF {#1} {\underline{f}} {
            \IfNoValueTF {#2} {\underline{f}^{(#1)}}{\underline{f}^{(#1)}_{#2}}
        }
}

\def\eqref#1{Eq.~(\ref{#1})}

\def\1{\bm{1}}

\def\vtheta{{\bm{\theta}}}
\def\vepsilon{{\bm{\epsilon}}}

\def\vdelta{{\bm{\delta}}}

\def\vx{{\bm{x}}}
\def\vy{{\bm{y}}}
\def\vz{{\bm{z}}}

\DeclareMathAlphabet{\mathsfit}{\encodingdefault}{\sfdefault}{m}{sl}
\SetMathAlphabet{\mathsfit}{bold}{\encodingdefault}{\sfdefault}{bx}{n}

\def\paperID{6974} 
\def\confName{CVPR}
\def\confYear{2024}

\usepackage[skip=6pt]{caption} 

\title{Can Protective Perturbation Safeguard Personal Data \\from Being Exploited by Stable Diffusion?}

\author{
  Zhengyue Zhao $^{1,2}$ \quad
  Jinhao Duan $^3$ \quad
  Kaidi Xu $^3$ \quad
  Chenan Wang $^3$\\
  Rui Zhang $^1$ \quad
  Zidong Du $^{1,4}$ \quad
  Qi Guo $^1$ \quad
  Xing Hu $^{1,4}$ $^\text{\Letter}$\\
  $^1$ SKL of Processors, Institute of Computing Technology, Chinese Academy of Sciences \\
  $^2$ University of Chinese Academy of Sciences \quad
  $^3$ Drexel University \\
  $^4$ Shanghai Innovation Center for Processor Technologies, SHIC \\
  \texttt{\small zhaozhengyue22@mails.ucas.ac.cn} \quad
  \texttt{\small \{jd3734, kx46, cw3344\}@drexel.edu} \\
  \texttt{\small \{zhangrui, duzidong, guoqi, huxing\}@ict.ac.cn}
}

\begin{document}
\maketitle

\begin{abstract}
Stable Diffusion has established itself as a foundation model in generative AI artistic applications, receiving widespread research and application. Some recent fine-tuning methods have made it feasible for individuals to implant personalized concepts onto the basic Stable Diffusion model with minimal computational costs on small datasets. However, these innovations have also given rise to issues like facial privacy forgery and artistic copyright infringement. In recent studies, researchers have explored the addition of imperceptible adversarial perturbations to images to prevent potential unauthorized exploitation and infringements when personal data is used for fine-tuning Stable Diffusion. Although these studies have demonstrated the ability to protect images, it is essential to consider that these methods may not be entirely applicable in real-world scenarios. In this paper, we systematically evaluate the use of perturbations to protect images within a practical threat model. The results suggest that these approaches may not be sufficient to safeguard image privacy and copyright effectively. Furthermore, we introduce a purification method capable of removing protected perturbations while preserving the original image structure to the greatest extent possible. Experiments reveal that Stable Diffusion can effectively learn from purified images over all protective methods\footnote{The code is available at \url{https://github.com/ZhengyueZhao/GrIDPure}}.

\end{abstract}    
\vspace{-10pt}
\section{Introduction}
\label{sec:intro}

\begin{figure}[t]
  \centering
   \includegraphics[width=0.95\linewidth]{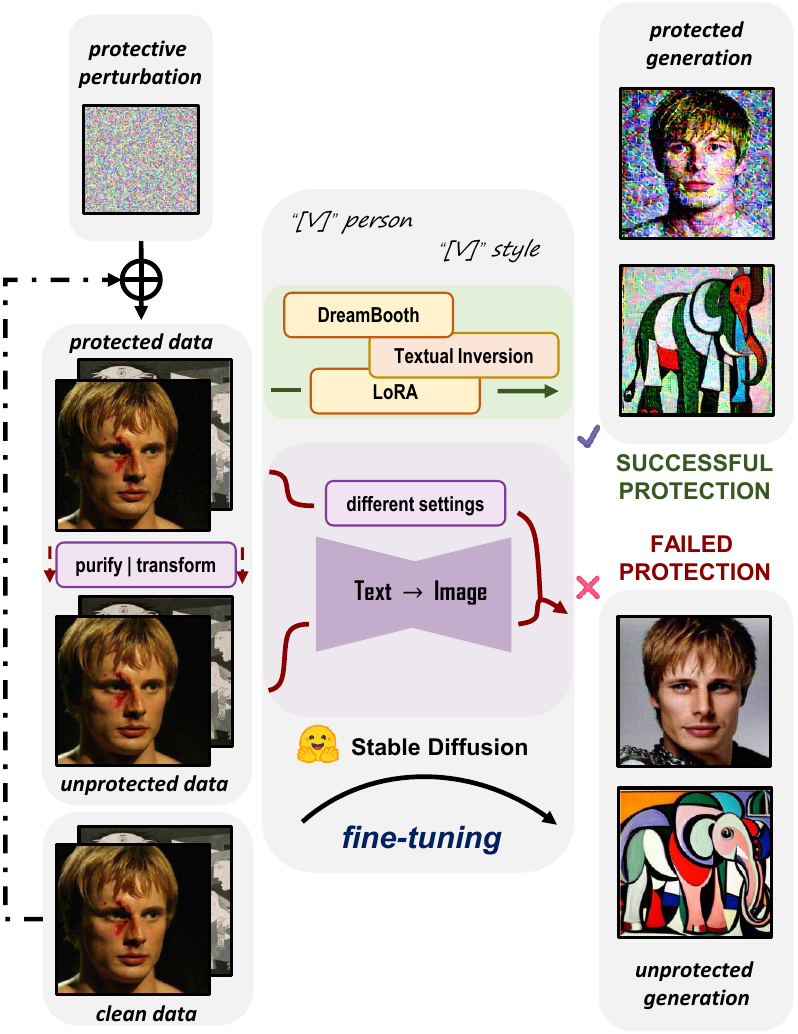}
   \caption{Overview of protective perturbation and failed protection facing exploitation of Stable Diffusion models.}
   \label{fig:overview}
   \vspace{-15pt}
\end{figure}

In recent years, Diffusion Models have achieved outstanding success in different domains~\cite{ho2020denoising,Score_Based,yuan2023remind,liu2024sora}.
In particular, Stable Diffusion, a multi-modal generative model built upon the framework of the Latent Diffusion Model~\cite{rombach2022high}, has garnered remarkable achievements in  AI-powered artistic applications. Considering the high requirements for training datasets and computational resources when starting from scratch, most of today's Stable Diffusion models are typically fine-tuned on top of a larger base model. Recently, fine-tuning methods such as Textual Inversion~\cite{gal2022image}, DreamBooth~\cite{ruiz2023dreambooth}, and Custom Diffusion~\cite{kumari2023multi} enable individual users to inject personalized concepts into the base model with minimal data and computational resources. These concepts can include specific individuals, objects, and unique styles. However, as Stable Diffusion gains widespread usage, concerns have emerged regarding image privacy and copyright issues. Fine-tuning on specific face datasets, for instance, allows Stable Diffusion to generate highly convincing images of individuals, leading to significant privacy breaches and authenticity concerns. Similarly, fine-tuning on the works of specific artists enables Stable Diffusion to easily replicate the artistic styles of these artists, potentially resulting in copyright infringement issues. These image privacy and copyright concerns raised by Stable Diffusion have attracted attention from society~\cite{bbc2022Art, cnn2022AI, washington2022AI}.

A series of research endeavors have been directed toward addressing image privacy and copyright issues introduced by Stable Diffusion~\cite{pmlr-v202-duan23b, SalmanKLIM23, gandikota2023erasing,kong2023efficient}. One notably prominent approach involves the addition of imperceptible protective adversarial perturbations to images, preventing Stable Diffusion from learning the features of protected images~\cite{pmlr-v202-liang23g, van2023anti, zheng2023understanding, shawn2023glaze, wu2023towards, ye2023duaw}. These efforts have showcased impressive results in safeguarding image data from being exploited by Stable Diffusion. After fine-tuning on images with adversarial perturbations, images generated by Stable Diffusion tend to exhibit lower quality and semantic deviations compared to results obtained from fine-tuning on clean images.
While these methods can ideally prevent Stable Diffusion from learning protected images, it's crucial to consider their effectiveness in more realistic scenarios. If these methods fail to adapt to various real-world usage contexts, they might give users a false sense of security~\cite{Radiya-Dixit0CT22}. Therefore, we need to subject this series of methods to a more realistic and systematic evaluation.

In this paper, we systematically examine the real-world application of safeguarding images from Stable Diffusion mining through adversarial perturbations, considering two main applications: protecting Stable Diffusion from learning the FaceID of a person and learning the style of an artist from artworks. Our examination includes various fine-tuning approaches for Stable Diffusion, diverse training data scenarios, and potential image transformations on the Internet. Our experiments indicate that it is difficult for the protective perturbation to safeguard personal images from being learned by Stable Diffusion under some complex practical conditions. We then explore defense mechanisms against these protective perturbations. We introduce \textbf{Gr}id \textbf{I}terative \textbf{D}iffusion-based \textbf{Puri}fication (\textbf{GrIDPure}), an extension of DiffPure, enabling effective purification of high-resolution images while preserving most of the structure of original images. Our results indicate that the method of protecting images through adversarial perturbations may not provide highly effective protection for personal images such as faces and image copyrights, which inspires us to seek more effective methods to prevent image copyright issues caused by generative AI.
Our contributions can be summarized as follows:
\begin{itemize}
\item We propose a practical threat model and meanwhile an applicable framework to comprehensively assess the effectiveness of privacy protection methods in the complex real-world environment and systematically evaluate the performance of multiple protective perturbation methods under the practical condition.
\item We analyze both the vulnerability of stable diffusion and the robustness of protective perturbations. We consider both natural perturbations that may decrease the protective effectiveness and the state-of-the-art adversarial purification model that can break the protection.
\item We propose GrIDPure, a simple yet effective purification method to remove adversarial perturbation from protected images and maintain the structure of the image. Results show that our method can effectively help Stable Diffusion learn the protected images.
\end{itemize}

\section{Background \& Related Works}
\label{sec:background}
\paragraph{Stable Diffusion.}
Stable Diffusion is based on the Latent Diffusion Model~\cite{rombach2022high}, which transfers diffusion models from pixel space to latent space with an image encoder and decoder. By introducing cross-attention layers into the UNet architecture, Stable Diffusion is able to generate high-resolution images with general conditional inputs.
\vspace{-10pt}
\paragraph{Fine-tuning Stable Diffusion.}
Considering the substantial computational requirements for training Stable Diffusion from scratch, many methods aim to inject specific concepts into Stable Diffusion through fine-tuning on base models. The fine-tuning methods currently in use include Textual Inversion~\cite{gal2022image}, DreamBooth~\cite{ruiz2023dreambooth}, Custom Diffusion~\cite{kumari2023multi}, and LoRA~\cite{hu2021lora}. Textual Inversion focuses solely on training a Text Embedding during the fine-tuning process to inject concepts into the text encoder without altering the weights of the UNet component. DreamBooth fine-tunes the entire UNet portion of the Stable Diffusion model. Unlike regular Text-to-Image fine-tuning, DreamBooth incorporates a prior loss during fine-tuning to prevent overfitting. Custom Diffusion identifies the cross-attention component in Stable Diffusion as the most crucial for the entire model, and it only modifies the weights of the cross-attention layer during fine-tuning. LoRA, on the other hand, trains weight increments in the attention layer of the UNet, enabling a quick and lightweight fine-tuning of Stable Diffusion. 
\vspace{-10pt}
\paragraph{Protective Perturbation against Stable Diffusion.}
To protect personal images such as faces and artwork from potential infringement when used for fine-tuning Stable Diffusion, recent research aims to disrupt the fine-tuning process by adding imperceptible protective noise to these images. Several methods have been developed to achieve this goal: Glaze~\cite{shawn2023glaze} focuses on preventing artists' work from being used for specific style mimicry in Stable Diffusion. It optimizes the distance between the original image and the target image at the feature level, causing Stable Diffusion to learn the wrong artistic style. AdvDM~\cite{pmlr-v202-liang23g} proposes a direct adversarial attack on Stable Diffusion by maximizing the Mean Squared Error loss during the optimization process. This approach uses adversarial noise to protect personal images. Anti-DreamBooth~\cite{van2023anti} incorporates the DreamBooth fine-tuning process of Stable Diffusion into its consideration. It designs a bi-level min-max optimization process to generate protective perturbations. Additionally, other research efforts~\cite{ye2023duaw, zheng2023understanding, wu2023towards, zhao2023unlearnable} have explored generating protective noise for images using similar adversarial perturbation methods.

\section{Threat Model}
\label{sec:threatmodel}
Considering that image infringement based on Stable Diffusion has practical implications, it is essential to define the threat model in real-world scenarios. We consider two participants involved in fine-tuning Stable Diffusion using images: the ``image protector'' and the ``image exploiter''. Specifically, we explain the workflow of the two parties as follows:

\noindent \textbf{Image Protector: } The Image Protector aims to provide protection for images to prevent exploitation by Stable Diffusion. In this context, the chosen protection method involves adding imperceptible protective perturbations to the images, with the goal of offering protection while minimizing alterations to the original image. In real-world scenarios, the Image Protector often faces challenges, such as not knowing the methods and forms the Image Exploiter will use to fine-tune Stable Diffusion with the protected images. Additionally, they cannot protect images that have been publicly disclosed in the past.

\noindent \textbf{Image Exploiter: } The Image Exploiter aims to fine-tune Stable Diffusion using images collected from the internet to generate high-quality images with specific concepts, including faces, objects, and artistic styles. To realistically assess the effectiveness of protective perturbations, we consider that the Image Exploiter may have the following possibilities during image collection and fine-tuning: (1) The Image Exploiter can choose any fine-tuning method, including but not limited to direct fine-tuning, LoRA, Textual Inversion, DreamBooth, and Custom Diffusion, among other mainstream fine-tuning methods. This requires the Image Protector to ensure that the protected images remain effective against any fine-tuning method. (2) During image collection, the Image Exploiter may gather both protected and unprotected images of the same concept (e.g., faces or styles). This necessitates the Image Protector to consider the effectiveness of protecting images with varying proportions among their publicly available images. (3) The protected images may undergo natural transformations during the dissemination process, including but not limited to cropping, compression, and blurring. This requires the Image Protector to consider the robustness of protective perturbations when exposed to these natural disturbances. (4) Image pre-processing: The Image Exploiter may employ purification methods to remove the protective perturbations from the collected images after acquisition.

We conduct a series of systematic evaluations of image protection methods using the more realistic threat model outlined above. This threat model can also serve as a fundamental framework for future researchers and users to assess methods for safeguarding image privacy.
\begin{table*}[t]
\small
\centering
\begin{tabular}{ccccccccccccc}
\toprule
\textbf{FT} & \multicolumn{3}{c}{\textbf{Text-to-Image (w\textbackslash te)}}& \multicolumn{3}{c}{\textbf{LoRA (w\textbackslash te)}}& \multicolumn{3}{c}{\textbf{DreamBooth (w\textbackslash te)}}& \multicolumn{3}{c}{\textbf{Textual Inversion}} \\
Metric & FID↓& CLIP↑ & prec. & FID↓& CLIP↑& prec. & FID↓& CLIP↑& prec.& FID↓& CLIP↑& prec.  \\ \hline
Clean  & 101.5& 0.7307& 0.80& 119.8  & 0.7378  & 0.64  & 95.90 & 0.7600 & 0.94 & 136.1  & 0.7881  & 0.3600 \\
AdvDM  & 240.4& 0.4419& 0.0 & 424.7  & 0.2316  & 0.0& 380.2 & 0.3500 & 0.0& 411.2  & 0.6539  & 0.0\\
AntiDB & 382.1& 0.3281& 0.0 & 439.1  & 0.2804  & 0.0& 408.4 & 0.3750 & 0.0& 500.6  & 0.5432  & 0.0\\
IAdvDM & 134.7& 0.7016& 0.86& 100.5  & 0.7028  & 0.82  & 174.0 & 0.5020 & 0.10 & 294.4  & 0.7226  & 0.02\\ \toprule
\textbf{FT} & \multicolumn{3}{c}{\textbf{Text-to-Image (w\textbackslash{}o te)}} & \multicolumn{3}{c}{\textbf{LoRA (w\textbackslash{}o te)}} & \multicolumn{3}{c}{\textbf{DreamBooth (w\textbackslash{}o te)}} & \multicolumn{3}{c}{\textbf{Custom Diffusion}}  \\
Metric & FID↓ & CLIP↑ & prec. & FID↓& CLIP↑& prec. & FID↓& CLIP↑& prec.& FID↓& CLIP↑& prec.  \\ \hline
Clean  & 155.3& 0.8284& 0.42& 157.2  & 0.8482  & 0.28  & 148.5 & 0.8352 & 0.54 & 139.8  & 0.8439  & 0.42\\
AdvDM  & 144.6& 0.7400& 0.56& 226.5  & 0.4868  & 0.08  & 173.2 & 0.6446 & 0.26 & 259.5  & 0.7471  & 0.0\\
AntiDB & 158.3& 0.5400& 0.32& 237.4  & 0.3726  & 0.10  & 215.3 & 0.3955 & 0.14 & 251.8  & 0.6641  & 0.0\\
IAdvDM & 146.3& 0.8484& 0.46& 134.1  & 0.7934  & 0.30  & 139.4 & 0.8708 & 0.58 & 156.6  & 0.8583  & 0.36\\ \bottomrule
\end{tabular}
\caption{Results of different protective perturbations toward different fine-tuning methods on the CelebA-HQ dataset. The first row reports results with training text encoder and the second row reports results without training text encoder.}
\label{tab:eval_ft}
\end{table*}

\vspace{-10pt}

\begin{table}[]
\small
\centering
\begin{tabular}{cccccc}
\toprule
\multicolumn{2}{c}{\multirow{2}{*}{\textbf{Fine-Tuning method}}} & \multicolumn{4}{c}{\textbf{Trainable Layers}}  \\ \cline{3-6} 
\multicolumn{2}{c}{} & VAE & Full-UNet & CA & TE \\ \hline
\multirow{2}{*}{Text-to-Image}  & w\textbackslash te& ×& $\surd$ & $\surd$ & $\surd$ \\ \cline{2-2}
  & w\textbackslash{}o te & ×& $\surd$ & $\surd$ & × \\ \cline{1-2}
\multirow{2}{*}{LoRA}& w\textbackslash te& ×& × & $\surd$ & $\surd$ \\ \cline{2-2}
  & w\textbackslash{}o te & ×& × & $\surd$ & ×  \\ \cline{1-2}
\multirow{2}{*}{DreamBooth} & w\textbackslash te& ×& $\surd$ & $\surd$ & $\surd$ \\ \cline{2-2}
  & w\textbackslash{}o te & ×& $\surd$ & $\surd$ & × \\ \cline{1-2}
\multicolumn{2}{c}{Textual Inversion}& ×& × & × & $\surd$\\ \cline{1-2}
\multicolumn{2}{c}{Custom Diffusion} & ×& × & $\surd$ & ×\\ \bottomrule
\end{tabular}
\caption{Comparison of different fine-tuning methods. CA represents cross-attention in the UNet and TE (te) represents text encoder. The text encoder can be chosen to be trained (w\textbackslash te) or fixed (w\textbackslash{}o te) in methods Text-to-Image, LoRA and DreamBooth.}
\label{tab:fine-tune_methods}
\vspace{-15pt}
\end{table}

\begin{figure}[t]
  \centering
\includegraphics[width=0.95\linewidth]{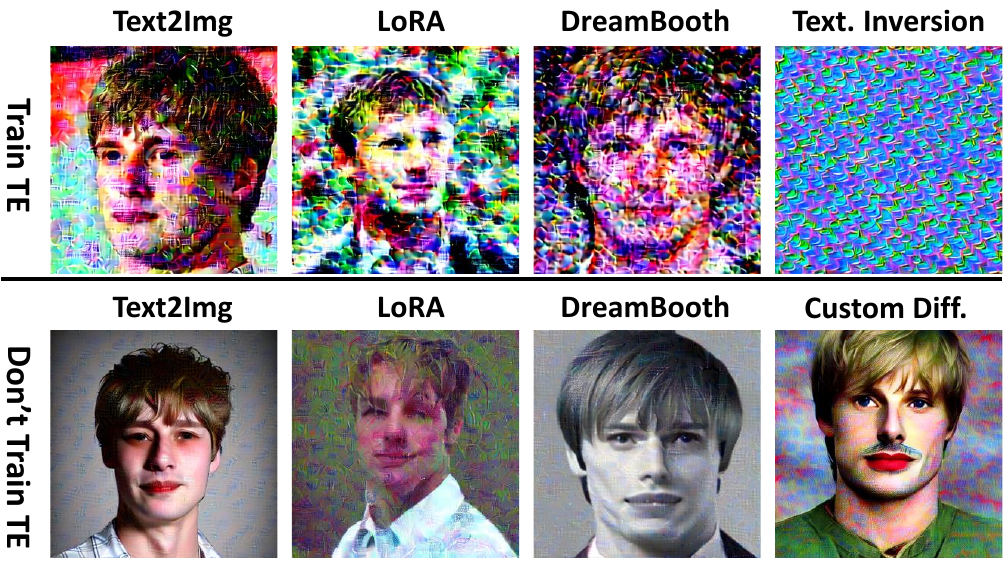}
\caption{Visualization of protective effectiveness of Anti-DreamBooth toward different fine-tuning methods on the CelebA-HQ dataset with prompt \textit{"a photo of a sks person"}.}
\label{fig:ft_celebahq}
\end{figure}
\begin{figure}[t]
  \centering
\includegraphics[width=0.95\linewidth]{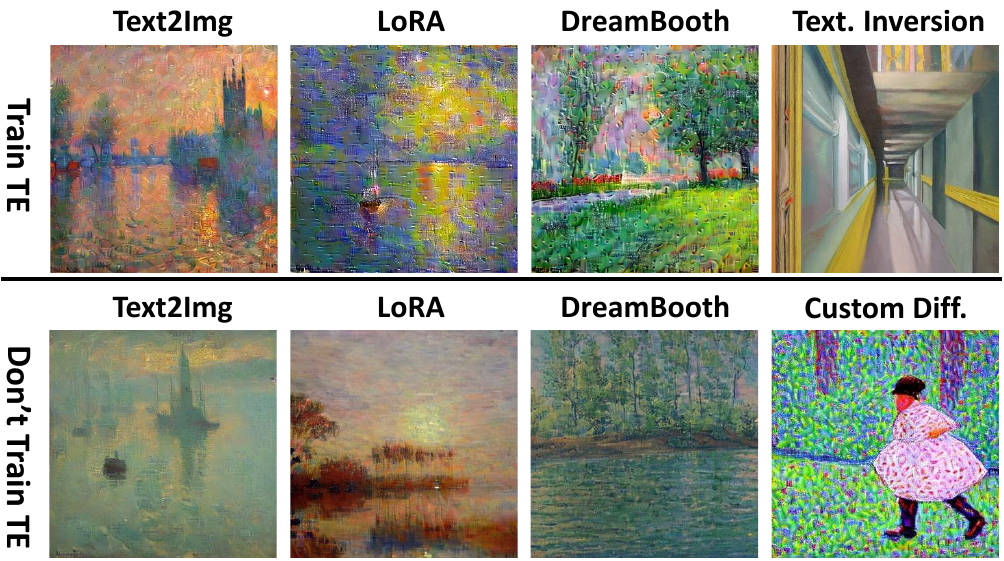}
\caption{Visualization of protective effectiveness of AdvDM toward different fine-tuning methods on the WikiArt dataset with prompt \textit{"a painting in the style of Monet"}.}
\label{fig:ft_wikiart}
\vspace{-15pt}
\end{figure}

\section{Evaluate the Protective Perturbation}
\label{sec:evaluate}
The evaluation focuses on two crucial applications: face protection and artwork style protection using high-quality datasets, CelebA-HQ~\cite{liu2015faceattributes}, and WikiArt~\cite{wikiart2016wikiart}, respectively. We choose AdvDM~\cite{pmlr-v202-liang23g} and DreamBooth~\cite{van2023anti} as the main protective perturbation methods in all the experiments which have better protective performance empirically. We also evaluate other perturbation methods such as Improved-AdvDM~\cite{zheng2023understanding} for face protection and Glaze~\cite{shawn2023glaze} for style protection in some experiments. We use two widely used image quality metrics, FID~\cite{heusel2017gans} and CLIP-Score~\cite{wang2023exploring}, to quantitatively demonstrate the generative quality, where lower FID and higher CLIP-Score represent better generative quality. Besides, we also provide the precision metric for generative models~\cite{kynkaanniemi2019improved} as a reference.

\subsection{Effectiveness Assessment in Fine-tuning}

\paragraph{Different Fine-Tuning Methods.}
To assess the effectiveness of these protective perturbations across various fine-tuning scenarios, we employ different fine-tuning methods for datasets with protection. As shown in Table~\ref{tab:fine-tune_methods}, these methods include direct Text-to-Image fine-tuning, LoRA, Textual Inversion, DreamBooth, and Custom Diffusion. For Text-to-Image, LoRA, and DreamBooth methods, they provide the option to train or not train the text encoder. In the case of Custom Diffusion, modifications are applied exclusively to the parameters in the key and value matrices of the cross-attention layers. The trainable layers within the Stable Diffusion models for each of these fine-tuning methods are outlined in the table. Our approach involves identifying suitable settings for each fine-tuning method using clean datasets initially. We then apply these settings to fine-tune Stable Diffusion models using protected datasets. This allows us to assess the impact of the protective perturbations across a range of fine-tuning approaches and quantitative results are shown in Table~\ref{tab:eval_ft}.

\begin{figure}[t]
  \centering
\includegraphics[width=0.95\linewidth]{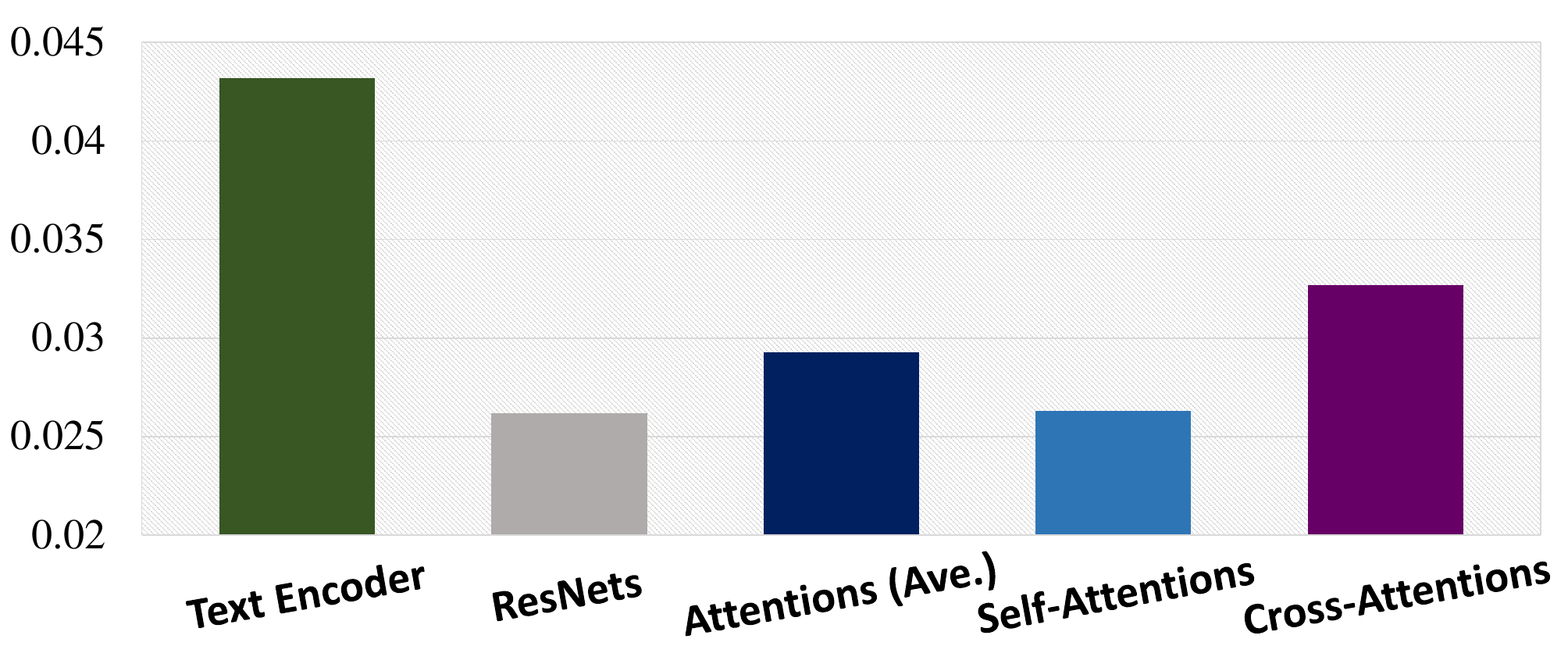}
\caption{Average changes of parameters ($\Delta\theta$) of different layers in the Stable Diffusion fine-tuned with clean and protected images.}
\label{fig:delta_theta}
\vspace{-15pt}
\end{figure}
\vspace{-10pt}
\paragraph{Text Encoder Makes Stable Diffusion More Vulnerable.}
To further exploit which layers of Stable Diffusion have more impact on the fine-tuning process, we design a similar experiment to what Custom Diffusion~\cite{kumari2023multi} does. Specifically, we calculate the relative difference between parameters of Stable Diffusion models fine-tuning with clean and perturbated images under the same initialization and training settings and the results are shown in Figure ~\ref{fig:delta_theta}.
\begin{equation}
\Delta\Bar{\vtheta}=\frac{1}{N}\sum_{n}{\frac{\Vert\vtheta_\text{adv}-\vtheta_\text{clean}\Vert}{\Vert\vtheta_\text{clean}\Vert}}
\end{equation}
This result indicates that the training of text encoder does make a great impact on image protection. It can be seen from results in Figure~\ref{fig:ft_celebahq} and ~\ref{fig:ft_wikiart} that, Textual Inversion, which fine-tunes the textual encoder only, shows the worst robustness toward protective perturbations and results in almost illegible generated images. Other fine-tuning methods that change parameters both in the UNet and text encoder also report worse generation quality compared with methods that only train the UNet. Table~\ref{tab:eval_ft} also supports the similar results that the text encoder is more vulnerable compared to other parts of Stable Diffusion models. 

\begin{figure}[t]
  \centering
\includegraphics[width=0.95\linewidth]{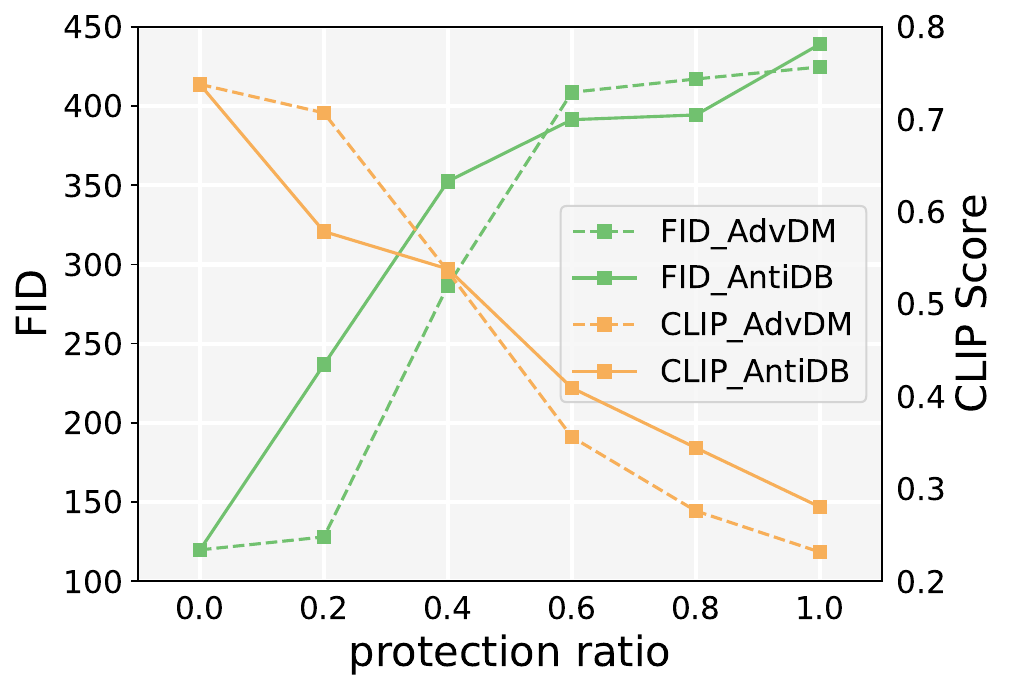}
\caption{Protective effectiveness of AdvDM and Anti-DreamBooth under different protective ratios of the CelebA-HQ dataset.}
\label{fig:fid_clip}
\vspace{-15pt}
\end{figure}
\vspace{-10pt}
\paragraph{Protection Ratio.}
We consider a common scenario in the practical training process: training a single concept with both clean and protected images. In this scenario, the image protector hopes to protect the image as strongly as possible even though some unprotected images have been released to the public. As a result, it's important to know the performance of protection with different protective ratios. We simulate different protection ratios from 0.2 to 1.0 to assess what unprotected images influence the protection effectiveness. Results in Figure~\ref{fig:fid_clip} indicate that both methods are sensitive to the protection ratio while AdvDM shows a worse protection performance when the protection ratio is small compared with Anti-DreamBooth.

\subsection{Natural Transformations Bypass Protection}
\begin{figure}[t]
  \centering
\includegraphics[width=0.98\linewidth]{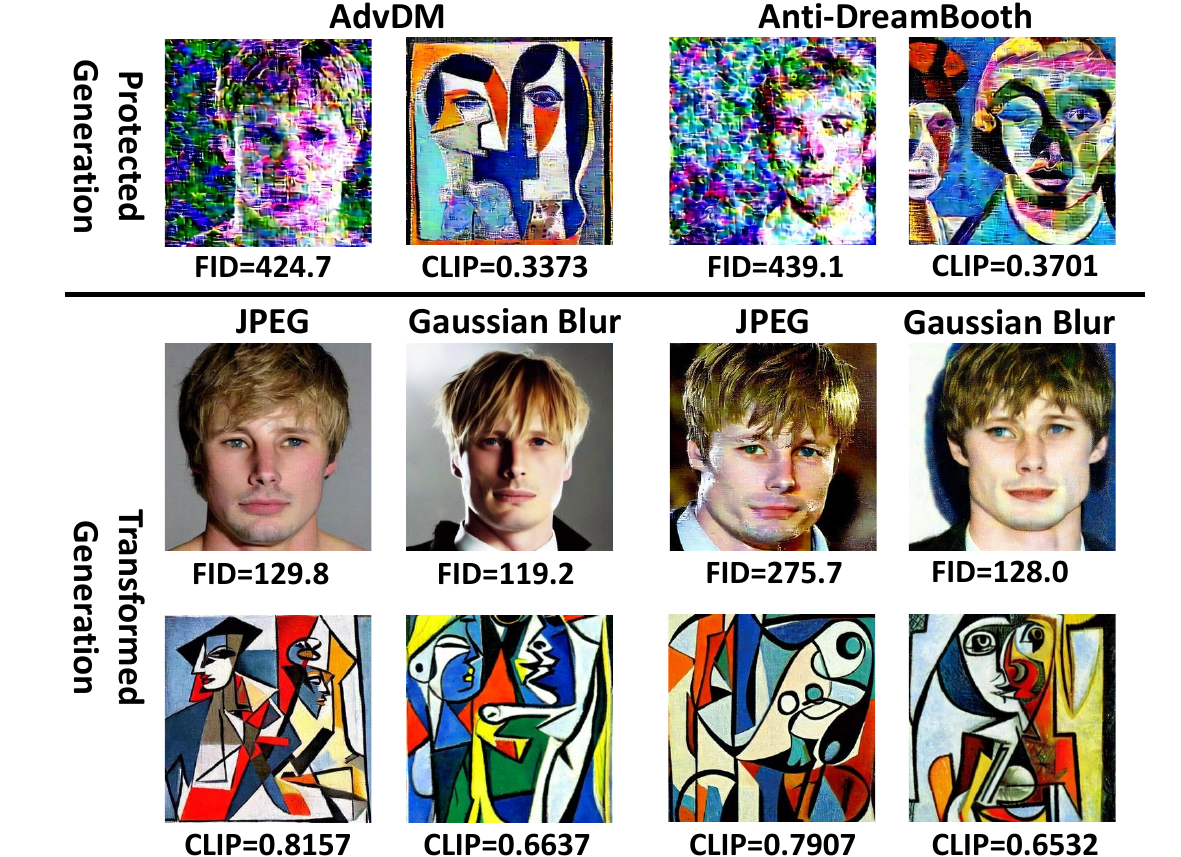}
\caption{Generated images of Stable Diffusion fine-tuned with LoRA on protected and transformed CelebA-HQ and WikiArt datasets. Natural transformation JPEG and Gaussian blur can significantly disrupt the protection of perturbations.}
\label{fig:natural_transform}
\end{figure}

Natural transformations such as compression and blur are common during the transmission of images on the internet. In this section, we assess the robustness of protective perturbations facing these natural transformations including JPEG compression and Gaussian blur with different strengthens.  We then adapt a classic robust-ascent algorithm Expectation over Transformation (EoT)~\cite{athalye2018synthesizing} to protective perturbations, to find out whether the protection can be more robust. Unfortunately, from the results in Figure~\ref{fig:natural_transform} we find that middle-strengthening natural transformations are strong enough to break the protection effectiveness of images. Though these natural transformations may decrease the quality and resolution of original images without doubt, these methods can still become image-preprocessing methods for image exploiters to bypass the protection with acceptable costs. 

\begin{equation}
\label{eq:diff-eot}
\max_{\Vert\vdelta_a\Vert<\rho} {\mathbb{E}_{\vepsilon, t, T} \Vert \vepsilon-\vepsilon_\vtheta(T(\mathcal{E}(\vx+\vdelta_a)), t) \Vert}^2
\end{equation}
We apply EoT to AdvDM as shown in Eq.~\ref{eq:diff-eot} with different transformations including color transformation, Gaussian blur, and so on. Results in Table~\ref{tab:eot} show that EoT doesn't help a lot when facing a middle-strengthening transformation, which indicates that it may be difficult to increase the robustness of protection toward natural transformations as pre-processing methods.
\begin{table}[]
\small
\begin{tabular}{ccccccc}
\toprule
\multirow{2}{*}{Dataset}& Trans. & No  & \multicolumn{2}{c}{JPEG} & \multicolumn{2}{c}{Blur} \\ \cline{2-2} \cline{4-7} 
  & Metric & Trans. & -  & +EoT& -  & +EoT\\ \hline
\multirow{2}{*}{CelebA} & FID↓& 424.7& 129.8  & 137.1  & 119.2  & 126.1  \\
  & CLIP↑  & 0.232  & 0.617 & 0.607 & 0.755 & 0.640 \\ \hline
\multirow{2}{*}{WikiArt}& FID↓& 251.1& 210.5  & 222.3  & 218.2  & 221.3  \\
  & CLIP↑  & 0.337  & 0.816 & 0.796 & 0.664 & 0.645 \\ \bottomrule
\end{tabular}
\caption{Robustness of protective perturbation optimized with Expectation over Transformation (EoT).}
\label{tab:eot}
\vspace{-15pt}
\end{table}

\section{Defense: GrIDPure}
\label{sec:defense}
\subsection{Purification Does Well}
\label{subsec:diffpure}
\begin{figure}
  \centering
\includegraphics[width=0.95\linewidth]{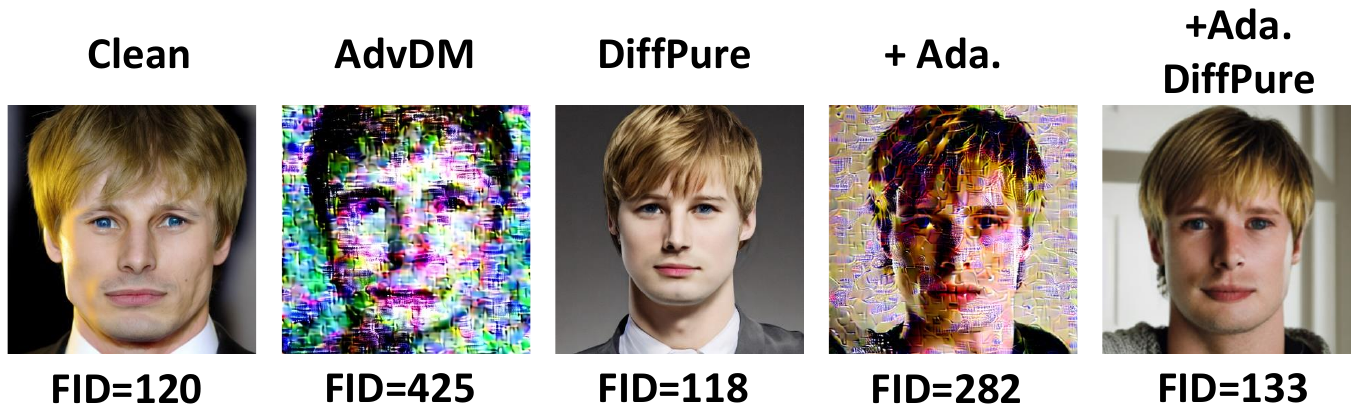}
\caption{Quantitative results and visualization examples of generated images of DiffPure towards AdvDM and adaptive attack.}
\label{fig:ada_diffpure}
\vspace{-15pt}
\end{figure}

It has been reported that adversarial purification can successfully remove adversarial perturbations from adversarial examples in classification tasks~\cite{pmlr-v139-yoon21a, Jiang2023UnlearnableEG, carlini2022certified, wang2022guided, pmlr-v162-nie22a}. Given that image exploiters might employ such techniques to purify protected images to bypass these protections after gathering them, it is crucial to assess the robustness of these protective perturbations against state-of-the-art purification methods.
\vspace{-10pt}
\paragraph{DiffPure.} DiffPure~\cite{pmlr-v162-nie22a} is the state-of-the-art adversarial purification method that utilizes SDEdit~\cite{meng2022sdedit} from an off-the-shelf unconditional diffusion model to purify adversarial images. In this process, an adversarial image is initially perturbed with Gaussian noise and subsequently denoised during the reverse steps of the diffusion models. To assess the robustness of the protection methods, we apply DiffPure to all the protective perturbations with various timesteps. The results in Table~\ref{tab:pure_effect} reveal that, with a sufficient number of forward steps, DiffPure can effectively recover the protected image to a learnable image.
\vspace{-10pt}
\paragraph{Adaptive Attack.} Some studies~\cite{Lee_2023_ICCV, kang2023diffattack, xue2023diffusionbased} have indicated that adaptive attacks can significantly reduce the effectiveness of DiffPure in classification tasks. If such adaptive settings can also undermine DiffPure in generative tasks, including Stable Diffusion fine-tuning, it would demonstrate the resilience of these protective measures against DiffPure. Consequently, we adopt the settings of Diff-PGD~\cite{xue2023diffusionbased} and full-gradient-based attacks used in image classification tasks to develop an adaptive attack against DiffPure.
\begin{equation}
\label{eq:diff-advdm}
 \max_{\Vert\vdelta_a\Vert<\rho} {\mathbb{E}_{\vepsilon, t} \Vert \vepsilon-\vepsilon_\vtheta(\text{denoise}(\text{diffusion}(\mathcal{E}(\vx+\vdelta_a), t_\text{pure})), t) \Vert}^2
\end{equation}
Specifically, we integrate the DiffPure process into the optimization of protective perturbation. This approach involved jointly computing the gradient of SDEdit and the loss of AdvDM, as shown in Eq.~\ref{eq:diff-advdm}. To ensure the computability of the gradient, we follow Diff-PGD, which utilizes DDIM~\cite{song2021denoising} to expedite the sampling during the reverse process. Our findings indicate that the protective perturbations generated with the adaptive attack exhibit a significant reduction in effectiveness when compared to the baseline perturbation. Unfortunately, as shown in Figure~\ref{fig:ada_diffpure}, these adaptive perturbations still fail to offer sufficiently robust protection against DiffPure. This may be attributed to the inherent instability of adversarial perturbations designed for generative models, which aim to shift the original distribution to another distribution, as opposed to those intended for classification models, which merely aim to alter the class label. This highlights the inadequacy of the perturbation's robustness against DiffPure, even with adaptive enhancements.

\begin{figure}
  \centering
\includegraphics[width=0.95\linewidth]{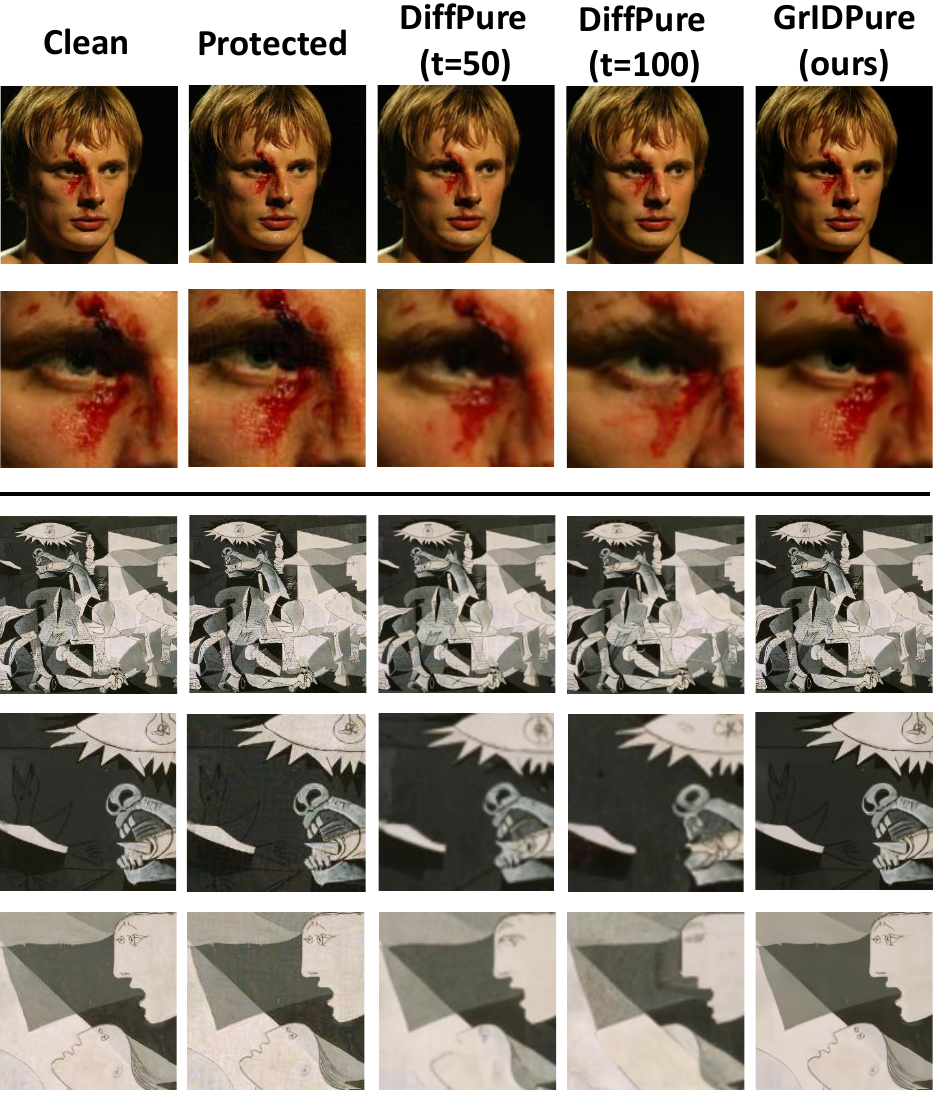}
\caption{Comparison of clean images, perturbated images and purified images. Our GrIDPure preserves most of the details of the original images (zoom in for better visualization).}
\label{fig:pure_quality}
\vspace{-10pt}
\end{figure}

\vspace{-10pt}
\paragraph{Limitaions of DiffPure.} While it's evident that DiffPure can effectively neutralize protective perturbations, its practical application for purging these perturbations faces challenges as visualized in Figure~\ref{fig:pure_quality}. Firstly, image exploiters seek purified images that closely resemble the original images. In a classification scenario, slight changes in an image may not significantly impact the final predicted label. However, for image generation tasks, preserving the intricate structures in images becomes paramount. This is particularly crucial for complex artworks with detailed elements, such as points and lines in an abstract painting by Picasso. Unfortunately, DiffPure with small timesteps falls short of completely eliminating the perturbation, while larger timesteps alter the image's structure, which may be unacceptable for high-quality and intricate artworks.

Additionally, the resolution of the purified image is closely tied to the diffusion model used for purification, typically limited to resolutions like $256\times256$ for models trained on ImageNet~\cite{deng2009imagenet, dhariwal2021diffusion}. This poses a significant limitation when applying DiffPure to Stable Diffusion Models, which demand high-resolution images for training or fine-tuning, often requiring resolutions like $512\times512$ and even more.

In response to these challenges, we introduce GrIDPure, an extension of DiffPure designed to retain the resolution and structure of the original image while effectively removing protective perturbations. This extension aims to bridge the gap and provide a more practical solution.
\begin{figure*}
  \centering
\includegraphics[width=0.95\linewidth]{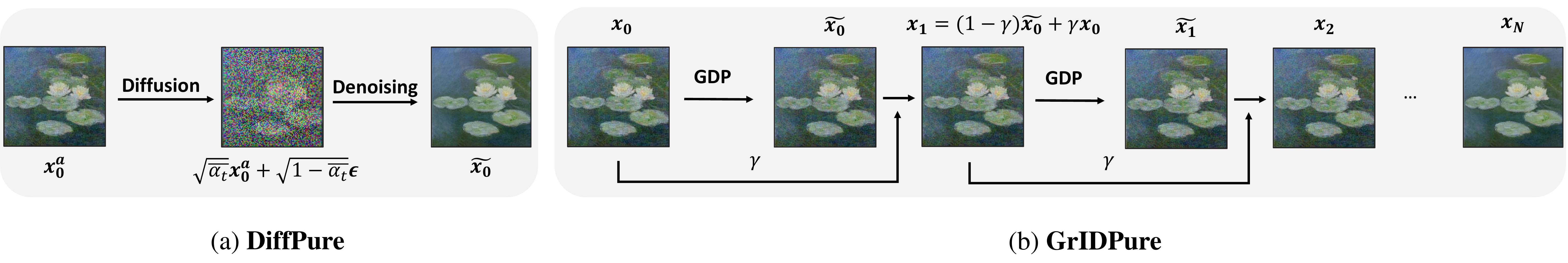}
\caption{Framework of (a) DiffPure and our proposed (b) GrIDPure.}
\label{fig:iter}
\end{figure*}

\begin{figure*}
  \centering
\includegraphics[width=0.9\linewidth]{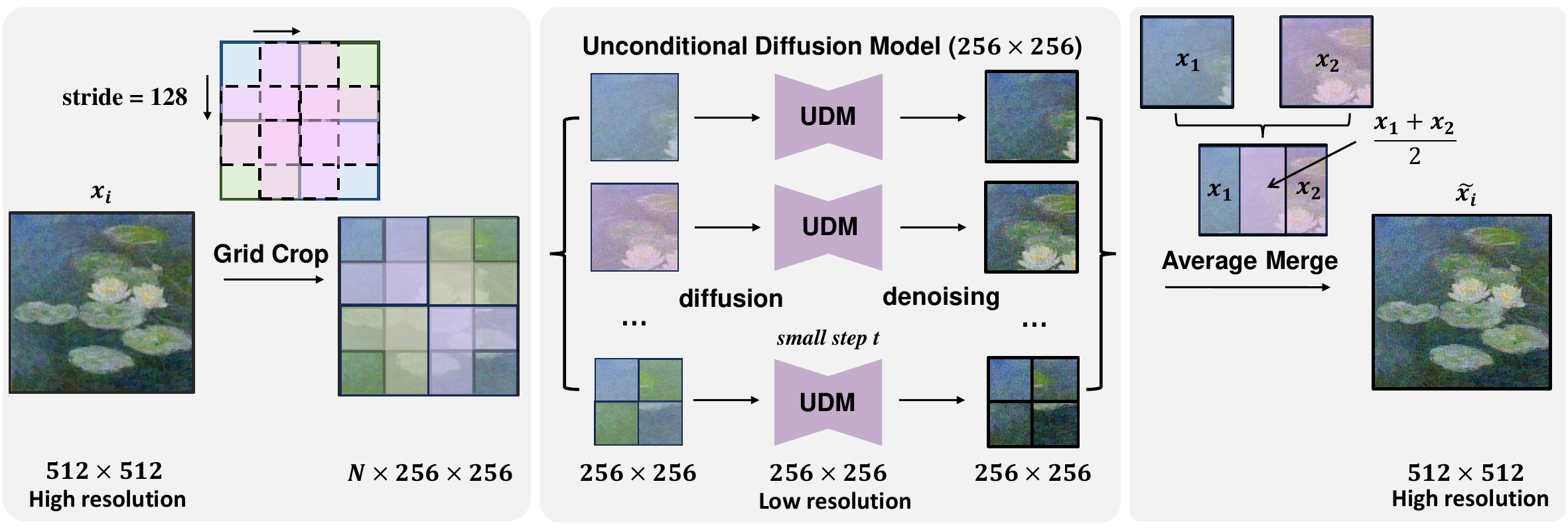}
\caption{The framework of grid diffusion-based purification (GDP). Each grid is diffused and denoised with a small step $t$. For example, given an image with resolution $512\times512$, we divide it into nine $256\times256$ grids with a 128-pixel overlap, ensuring that each pair of adjacent grids shares a region of $256\times128$ pixels. The four $128\times128$-pixel corners are combined into a $256\times256$ grid to ensure they are part of two different grids. Under these conditions, the image is ultimately divided into ten $256\times256$ grids.}
\label{fig:gdp}
\end{figure*}

\subsection{GrIDPure}
Our proposed \textbf{Gr}id \textbf{I}terative \textbf{D}iffusion-based \textbf{Puri}fication (\textbf{GrIDPure}) is a purification method designed to preserve image resolution and intricate details. Specifically, we introduce small-step iterative DiffPure to preserve the details and apply grid-based cropping to preserve the resolution of the image. The process involves several key steps: (1) The high-resolution image is initially divided into multiple grids, ensuring that each part of the image overlaps with at least two grids. (2) Each grid is then purified using SDEdit, employing an unconditional diffusion model with small steps. (3) The purified grids are merged back into a high-resolution image, with any overlapping parts being averaged during the merging process. (4) The merged image is blended with the original image, with the blending ratio controlled by a weight parameter $\gamma$. The entire process is iterated multiple times to produce the final purified image.
\vspace{-10pt}
\paragraph{Iterative DiffPure With Small Steps.} 
Considering that these imperceptible protective perturbations are usually high-frequency noise, removing this noise while preserving the original details of the image is possible through small-step SDEdit. Besides, iterating the small-step DiffPure multiple times can get better purification efficacy~\cite{Lee_2023_ICCV, pmlr-v139-yoon21a}. This insight us to break down a large-step DiffPure into a series of smaller-step DiffPure iterations. We empirically verify that iterative DiffPure with small steps can effectively remove protective noise (as shown in Figure~\ref{fig:IDP}) and better preserve image details compared to DiffPure (as shown in Figure~\ref{fig:ablation_iter_grid} and Table~\ref{tab:ablation_iter_grid}). 

\vspace{-10pt}
\paragraph{Grid Diffusion-based Purification.}

Cropping the image into several grids and then merging all the purified grids to create the final image can lead to difficulties in grid merging. As a solution, we apply a small-step purification to each grid instead of using the full DiffPure process, as shown in Figure~\ref{fig:gdp}. To be more specific, we divide the input image into several grids, ensuring that each grid has the same resolution as the unconditional diffusion model for purification. To eliminate any unnatural borders between the grids, we make sure that each part of the image overlaps with at least two different grids. 
\vspace{-10pt}
\paragraph{Average Merge.}
After each grid is purified with the small-step DiffPure method, all the grids are merged into an image with the same resolution as the original image. To manage the overlapped sections of each pair of nearby grids, we calculate the average of the shared parts across all overlapped grids. Through small-step purification, overlapped grid cropping, and average merging, we can effectively and seamlessly purify high-resolution images.
\vspace{-10pt}
\paragraph{Implementation of Iteration.}
Each iteration consists of grid diffusion-based purification and the blending operation of the purified image with the original image. More specifically, an image $x_i$ undergoes the above processing, resulting in $\Tilde{x_i}$. Subsequently, $\Tilde{x_i}$ is blended with $x_i$ using the blending weight $\gamma$:  $\vx_{i+1} = (1-\gamma)\cdot\Tilde{\vx}_i+\gamma\cdot \vx_i$.
The purpose of blending is to regulate the purification rate and contribute to the preservation of the original image's structure.

\subsection{Evaluation of GrIDPure}
We assess our GrIDPure approach in two stages. First, we compare the quality of images purified by DiffPure and our GrIDPure. Then, we demonstrate that our GrIDPure effectively eliminates protective perturbations added to protected images.
\vspace{-20pt}
\paragraph{Quality of Purification.} 
The quality of purified images is crucial in generative tasks, as it affects factors such as resolution and the preservation of the image's structure. We conduct both qualitative and quantitative comparisons between our GrIDPure and DiffPure. To do this, we utilize datasets of artworks from both CelebA-HQ and WikiArt datasets and evaluate the results in high-resolution ($512\times512$) applications. Quantitative assessment is performed using the Structure Similarity Index Measure (SSIM)~\cite{wang2004image} and Peak Signal-to-Noise Ratio (PSNR) to measure the similarity between the purified image and the original clean image. Higher SSIM and PSNR values indicate better purification quality. The results Figure~\ref{fig:pure_quality} and Table~\ref{tab:pure_quality} demonstrate that GrIDPure effectively preserves the resolution and most of the detailed structure of the original clean image.

\begin{table}[]
\small
\centering
\adjustbox{width=0.48\textwidth}{
\begin{tabular}{lccccc}
\toprule
\multirow{2}{*}{Dataset} & \multirow{2}{*}{Pert.} & \multirow{2}{*}{Metric} & DiffPure & DiffPure & \multirow{2}{*}{GrIDPure} \\
 & & & (t=50)& (t=100)  &\\ \hline
\multirow{4}{*}{CelebA}  & \multirow{2}{*}{AdvDM}  & FID↓  & 192.9 & \textbf{117.7} & 121.4\\
 & & CLIP↑ & 0.6834& 0.7400& \textbf{0.8526}  \\ \cline{2-6} 
 & \multirow{2}{*}{AntiDB} & FID↓  & 145.3 & 133.9 & \textbf{125.5}\\
 & & CLIP↑ & 0.7225& 0.7040& \textbf{0.7773}  \\ \bottomrule
\multirow{6}{*}{WikiArt} & \multirow{2}{*}{AdvDM}  & FID↓  & 203.5 & 214.2 & \textbf{203.4}\\
 & & CLIP↑ & 0.7449& 0.7843& \textbf{0.8758}  \\ \cline{2-6} 
 & \multirow{2}{*}{AntiDB} & FID↓  & \textbf{196.4} & 200.4 & 197.3\\
 & & CLIP↑ & \textbf{0.7824}& 0.7401& 0.7818  \\ \cline{2-6} 
 & \multirow{2}{*}{Glaze} & FID↓  & 200.6 & 202.9 & \textbf{195.3} \\& & CLIP↑ & 0.7410&0.7320 &\textbf{0.7981}\\
 \bottomrule
\end{tabular}
}
\caption{Generative quality of Stable Diffusion fine-tuned with purified datasets. See Appendix~\ref{sec:results} for more resuls. }
\label{tab:pure_effect}
\vspace{-10pt}
\end{table}

\begin{table}[]
\small
\centering
\begin{tabular}{ccccc}
\toprule
\multirow{2}{*}{Pert.} & \multirow{2}{*}{Metric} & DiffPure & DiffPure & \multirow{2}{*}{GrIDPure} \\
& & (t=50)& (t=100)  &\\ \hline
\multirow{2}{*}{AdvDM} & PSNR↑ & 23.20 & 22.24 & \textbf{30.60}\\
& SSIM↑ & 0.6978& 0.6378& \textbf{0.9199}  \\ \hline
\multirow{2}{*}{AntiDB} & PSNR↑ & 23.16 & 22.19 & \textbf{30.63}\\
& SSIM↑ & 0.6858& 0.6342& \textbf{0.9156}  \\ \bottomrule
\end{tabular}
\caption{Quantitative results of the average purification quality of DiffPure and GrIDPure. Images purified via GrIDPure are more similar to the original images.}
\label{tab:pure_quality}
\vspace{-20pt}
\end{table}

\vspace{-15pt}
\paragraph{Effectiveness of Purification.}
We assess the efficacy of GrIDPure against multiple protective methods, including Glaze, AdvDM and Anti-DreamBooth. Our evaluation involves examining the images generated by Stable Diffusion models fine-tuned on purified datasets for both face generation and artwork style mimicry applications. The results in Table~\ref{tab:pure_effect} indicate that images purified through GrIDPure can be successfully learned by Stable Diffusion, and the fine-tuned model can generate high-quality images. This suggests that GrIDPure offers a more versatile and adaptable approach for generative tasks to bypass these protective perturbations, rendering the protection ineffective.
\vspace{-15pt}
\paragraph{Limitations.} GrIDPure exhibits a higher time complexity compared to other purification methods due to the grid cropping operation. Our experiments reveal that it takes approximately 2 minutes to purify a $512\times512$ image on a single V100 GPU using our default settings. However, this time investment is not a significant concern, especially given that the datasets for fine-tuning Stable Diffusion are typically small in size. In contrast, protective methods, such as Glaze, can be even more time-consuming, taking about 10 minutes per image. It's worth noting that the workflow of GrIDPure is amenable to parallel acceleration. Additionally, GrIDPure perseveres original image quality better than  DiffPure, purification methods are more suited for photos and modern-style artworks than oil paintings though. This is because purification methods may inadvertently remove some oil texture, which is typically intertwined with protective perturbations in oil paintings.

\section{Conclusion}
\label{sec:conclusion}
In this paper, we provide a systematic and realistic discussion of the method using adversarial perturbations to protect image data from unauthorized exploitation by Stable Diffusion. Our experimental results suggest that, in practical applications, this protection method is fragile and unreliable, primarily due to the following reasons: 

\begin{itemize}
\item Firstly, the effectiveness of perturbation-based protection varies significantly depending on different fine-tuning methods. These perturbations rely on attacks against the text encoder and yield smaller benefits for methods that do not require fine-tuning of the text encoder. This makes protections unstable considering that image protectors cannot determine the fine-tuning methods employed by image exploiters.

\item Secondly, this protection method is sensitive to the proportion of images being protected. In real-world applications, image protectors cannot protect images that have already been publicly shared. This means that image miners may collect both protected and unprotected images while gathering data. Perturbing images in this manner is insufficient to provide effective protection when the proportion of protected images is small.

\item Thirdly, this protection method lacks robustness. Natural transformations such as Gaussian blur and JPEG can significantly reduce the protection effectiveness. These transformations are common during internet transmission and image pre-processing. Some methods designed to enhance robustness, like Expectation over Transformation (EoT), also struggle to provide security for such image generation tasks.

\end{itemize}

Finally, we propose an effective method to remove these adversarial perturbations, GrIDPure, an extension of DiffPure. It effectively removes adversarial perturbations while better \textbf{preserving the original features} of the image, allowing Stable Diffusion to learn semantics closer to clean images when trained on purified images.

{
    \small
    \bibliographystyle{ieeenat_fullname}
    \bibliography{main}
}

\clearpage
\maketitlesupplementary
\setcounter{section}{0}
\renewcommand{\thesection}{\Alph{section}}

\section{Preliminary and Implementations}
\label{sec:implementation}
\subsection{Fine-tuning Stable Diffusion}

\paragraph{Text-to-Image.}
To directly fine-tune a Stable Diffusion model, users need to optimize the following loss function:
\begin{equation}
 \label{equ:ldm_loss}
\mathcal{L}_{\text{LDM}}=\mathbb{E}_{t,\vx_0,\vepsilon}\left[{\Vert \vepsilon - \vepsilon_\vtheta(\vz_t, t, \tau_\vtheta(\vy)) \Vert}^2 \right]
\end{equation}
Where $\vz_t$ is the latent vector generated by the image in pixel space $\vx_0$ and an image encoder $\mathcal{E}(\cdot)$. $\vy$ is the text embedding and $\tau_\vtheta(\cdot)$ is the layers in Stable Diffusion which align the text embedding with the latent image vector.

\paragraph{LoRA.}
Low-Rank Adaptation (LoRA)~\cite{hu2021lora} is a light fine-tuning method designed for large language models, which introduces rank decomposition matrices of Transformer layers to make the fine-tuning process more efficient, as shown in Eq.~\ref{eq:lora}. $W_0$ is a pre-trained weight matrix and $B$ and $A$ are low rand decomposition matrices of $\Delta W$.
\begin{equation}
    \label{eq:lora}
    h = W_0 \vx + \Delta W \vx = W_0 x + BA\vx
\end{equation}
Ryu et al. introduce LoRA into Stable Diffusion for fast text-to-image diffusion fine-tuning \footnote{\url{https://github.com/cloneofsimo/lora}}, providing an efficient training and small size outputs for Stable Diffusion fine-tuning.

\paragraph{DremBooth.}
DreamBooth~\cite{ruiz2023dreambooth} combines the reconstruction loss of diffusion training with a class-specific prior preservation loss to better avoid overfitting when fine-tuning Stable Diffusion with just several images.
\begin{equation}
    \label{eq:dreambooth}
    \begin{aligned}
        \mathbb{E}_{x,c,\epsilon,\epsilon ', t} [\omega_{t}\Vert\hat{\vx}_\theta(\alpha_{t}\vx+\sigma_t\vepsilon, c)\Vert^2_2 +\\
        \lambda\omega_{t'}\Vert\hat{\vx}_\theta(\alpha_{t'}\vx_{pr}+\sigma_{t'}\vepsilon ', c_{pr})\Vert^2_2]
    \end{aligned}
\end{equation}
The second term of the training loss in Eq.~\ref{eq:dreambooth} is the prior-preservation loss which supervises the Stable Diffusion with its class-specific generated images.

\paragraph{Custom Diffusion.}
Custom Diffusion~\cite{kumari2023multi} updates weights in Key and Value matrices of cross-attention layers while freezing other layers in the Stable Diffusion model, which are more influential during the text-to-image fine-tuning. Besides, Custom Diffusion also uses a regularization set of real images to prevent overfitting and use text encoding to better inject the new concept. In our implementation, we only train the cross-attention layer and keep the weights of the text encoder during fine-tuning to highlight the features of Custom Diffusion.

\paragraph{Textual Inversion.} 
Textual Inversion~\cite{gal2022image} finds a target token $v_*$ to match the personal concept by directly optimizing the LDM object as shown in Eq.~\ref{eq:textinversion}. 
\begin{equation}
    \label{eq:textinversion}
    v_*=\arg\min_v \mathbb{E}_{\vz,\vy,\vepsilon,t}[\Vert\vepsilon-\vepsilon_\vtheta(\vz_t,t,c_\vtheta(\vy))\Vert^2_2]
\end{equation}
The advantage of Textual Inversion compared with other fine-tuning methods is that it only changes the text encoder of the Stable Diffusion model and keeps all parameters in the UNet while fine-tuning, which enables users to inject personal concepts with much smaller computational and spatial overhead.

\subsection{Protective Perturbations}

\paragraph{AdvDM.}
AdvDM~\cite{pmlr-v202-liang23g} introduces a simple yet effective pipeline to add $l_\infty$ adversarial perturbations into images. The basic motivation of AdvDM is to make generative images be out-of-distribution examples, which leads to maximizing the following object in Eq.~\ref{eq:advDM}:
\begin{equation}
\label{eq:advDM}
\max_{\Vert\vdelta_a\Vert<\rho} {\mathbb{E}_{\vepsilon, t} \Vert \vepsilon-\vepsilon_\vtheta(\mathcal{E}(\vx+\vdelta_a), t) \Vert}^2
\end{equation}
Where $\rho$ is the $l_\infty$ bound of the perturbations. Its results show that images protected by AdvDM can be prevented from being used for style transfer and Stable Diffusion fine-tuning.

\paragraph{Anti-DreamBooth.}
Anti-DreamBooth~\cite{van2023anti} proposes another strong method to optimize the protective perturbation. Specifically, Anti-DreamBooth alternatively optimizes perturbations by maximizing the training loss of LDM and minimizing the training loss of DreamBooth to change the parameters of the model, as shown in Eq.~\ref{eq:anti-dreambooth}:
\begin{equation}
    \label{eq:anti-dreambooth}
    \begin{aligned}
    \vdelta=\arg\max_{\Vert\vdelta\Vert_p<\rho}\mathcal{L}_{\text{LDM}}(\vtheta,\vx) \\
    \vtheta=\arg\min_{\vtheta}\mathcal{L}_{\text{DreamBooth}}(\vtheta,\vx+\vdelta)
    \end{aligned}
\end{equation}
Although it seems that Anti-DreamBooth is designed for the DreamBooth fine-tuning method, our experiments indicate that Anti-DreamBooth is also effective in other fine-tuning methods both in face and style learning.

\paragraph{Glaze.}
Glaze~\cite{shawn2023glaze} focuses on the copyright concerns of style mimicry of text-to-image models. Different from the full-model attack in AdvDM and Anti-DreamBooth, Glaze designs a targeted optimization object toward the image feature-extracting process, which corresponds to the VAE in the Stable Diffusion model. 
\begin{equation}
    \label{eq:glaze}
    \min_\vdelta\Vert\Phi(\Omega(\vx,T)),\Phi(\vx+\vdelta)\Vert^2_2+\alpha\cdot\max(\text{LPIPS}(\vdelta)-p,0)
\end{equation}
As shown in Eq.\ref{eq:glaze}, $\Phi(\cdot)$ is the image feature extractor, $\Omega(\cdot)$ refers to the style transfer, while $T$ is the targeted style. Following this pipeline, Glaze aims to make Stable Diffusion learn the targeted style instead of the real style of training images during the fine-tuning process.

\subsection{Expectation over Transformation}
Expectation over Transformation (EoT)~\cite{athalye2018synthesizing} is firstly proposed to synthesize physical adversarial examples in the real-world environment, which is an effective robust-ascending method towards tons of physical transformation such as cropping, rotation and color transformations. To evaluate the robustness of protective perturbation on Stable Diffusion models, we adopt the EoT methods to AdvDM (as shown in Eq.~\ref{eq:diff-eot}) to assess whether EoT helps to defend natural image transformations such as compression and blur. Specifically, we sample transformations of EoT including regular color transformations and Gaussian blur, which are usually used in the traditional EoT on classification tasks.

\subsection{Adaptive Attack against DiffPure}
To further evaluate the robustness of adversarial perturbations when facing the state-of-the-art adversarial purification method, DiffPure~\cite{pmlr-v162-nie22a}, we design an adaptive attack pipeline following recent diffusion-based attack methods for classification tasks~\cite{kang2023diffattack, xue2023diffusionbased}. More specifically, we adopt the implementation of~\cite{xue2023diffusionbased} which alternates the reverse sampling of SDEdit into DDIM to speed the back-propagation while also preventing memory overflow of GPUs, which is represented in Eq.~\ref{eq:diff-advdm}. 
\begin{figure}[htbp]
  \centering
   \includegraphics[width=0.98\linewidth]{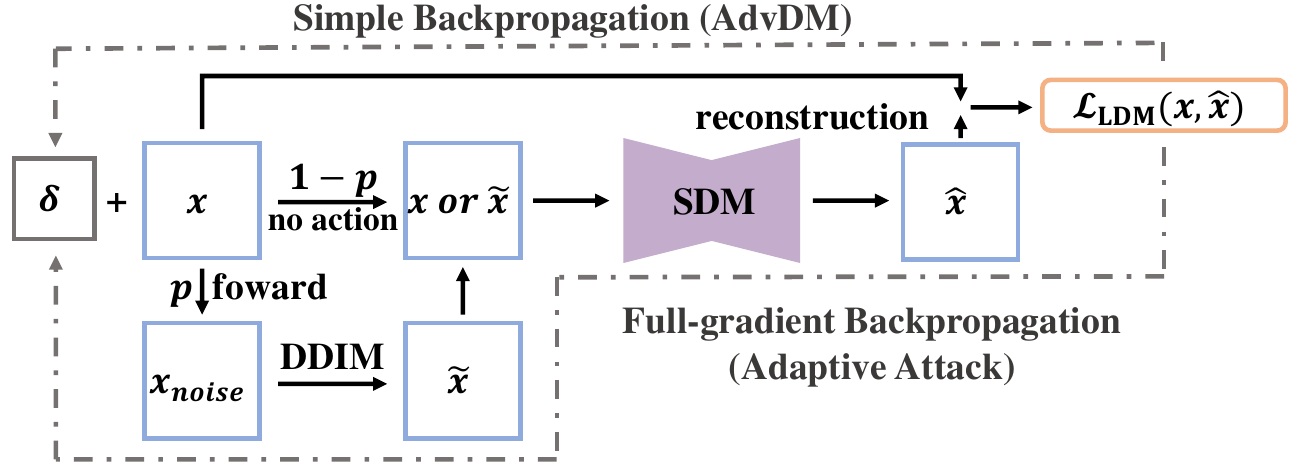}
   \caption{Framework of adaptive attack against DiffPure.}
   \label{fig:ada_diffpure_framwork}
\end{figure}
However, we find that directly applying the full-gradient back-propagation to the optimization of perturbations will lead to failed protection even without purification, which may mainly be due to the Monte Carlo sampling of timestep $t$ and noise $\vepsilon$. Thus, we make a trade-off between the simple and the full-gradient back-propagation under the control of a probability $p$, as shown in Figure~\ref{fig:ada_diffpure_framwork}.
By searching for the value of $p$, we find that setting $p$ to $0.2$ can get the best results in both with and without purification scenarios.

\subsection{GrIDPure}
Our GrIDPure is an iterative purification method as demonstrated in Figure~\ref{fig:iter} and Algorithm~\ref{alg:GrIDPure}, which contains several GDP iterations (Figure~\ref{fig:gdp} and Algorithm~\ref{alg:GDP}). There are two crucial parameters to control the purification pipeline, including the number of iterations $M$ and the number of purification steps $T$ of each iteration. Larger $M$ and $T$ can help to purify adversarial perturbations more completely but also lead to less similarity compared with the original clean image.

\begin{CJK}{UTF8}{gbsn} 
\begin{algorithm} [htbp]
\caption{GrIDPure}
\label{alg:GrIDPure}
    \floatname{algorithm}{Algorithm}
    \begin{algorithmic}[1]
    \Require Perturbated Image $\vx^{a}$, Blend Weights $\gamma$, Purification steps $T$, Iterations $M$, the number of grids $K$
    \Ensure Purified Image $\vx^{p}$
        \State Initialize $ \vx_0 \leftarrow \vx^{a}$
        \For{$m = 0 \rightarrow M-1$}
            \State $\tilde{\vx}_m \leftarrow$ \textbf{GDP}($\vx_m, T, K$)
            \State $\vx_{m+1} \leftarrow (1-\gamma) \cdot \tilde{\vx}_m + \gamma \cdot \vx_m$
        \EndFor
        \State $\vx^{p}\leftarrow \vx_M$
        \State \Return $\vx^{p}$
    \end{algorithmic} 
\end{algorithm}
\end{CJK}

\begin{CJK}{UTF8}{gbsn} 
\begin{algorithm} [htbp]
\caption{Grid Diffusion-based Purification (GDP)}
\label{alg:GDP}
    \floatname{algorithm}{Algorithm}
    \begin{algorithmic}[1]
    \Require Perturbated Image $\vx$, the number of grids $K$, Purification steps $T$
    \Ensure Purified Image $\tilde{\vx}$
        \State $\vx^{0}, \vx^{1}, ..., \vx^{K-1} \leftarrow$ \textbf{Crop}($\vx, K$)
        \For{$k = 0 \rightarrow K-1$}
            \State $\vx^k_{\text{n}} \leftarrow \text{diffusion}(\vx^k, T)$
            \State $\tilde{\vx}^k \leftarrow \text{denoise}(\vx^k_{\text{n}}, T)$
        \EndFor
        \State $\tilde{\vx} \leftarrow \textbf{Merge}(\tilde{\vx}^0, ..., \tilde{\vx}^{K-1})$
        \State \Return $\tilde{\vx}$
    \end{algorithmic} 
\end{algorithm}
\end{CJK}

\section{Experiments Settings}
\label{sec:settings}
\subsection{Datasets and Metrics}
We run the experiments on two main datasets: CelebA-HQ and WikiArt. CelebA-HQ is a high-quality dataset with a resolution of $1024\times1024$ that contains over 15 images for each attribute. WikiArt is an open-source painting dataset that contains artworks of different artists, we choose 6 to 10 images per artist to simulate usual practical fine-tuning. We resize the resolution of all these images to $512\times512$ to match the images with the base Stable Diffusion model (\textit{stable-diffusion-v1.5}\footnote{\url{https://huggingface.co/runwayml/stable-diffusion-v1-5}}). To assess the generative quality, we generate 100 images for each concept and calculate two full-reference indexes, FID and precision score, and a non-reference quality metric, CLIP-Score. The FID and precision are based on the evaluation of \textit{guided diffusion}\footnote{\url{https://github.com/openai/guided-diffusion}}, while the CLIP-Score is based on the CLIP-IQA of the \textit{piq} library\footnote{\url{https://github.com/photosynthesis-team/piq}}. In the experiment of purification quality which needs to compare the similarity between the purified image and the original clean image, we use SSIM and PSNR to evaluate the purification. Both of the indexes are based on the \textit{piq} library.
\subsection{Settings of Fine-tuning Methods}
We choose LoRA as the default fine-tuning method in all the experiments, which is one of the most popular methods in the AIGC community. Besides, LoRA with training Text Encoder is also one of the most vulnerable fine-tuning methods in our results, which can help us further explore whether the protection is valid or not. For experiments in evaluating different fine-tuning methods (in Section~\ref{sec:evaluate}), we apply all 8 different fine-tuning methods as shown in Table~\ref{tab:fine-tune_methods}. For detailed parameters of fine-tuning, the learning rates of Text-to-Image, DreamBooth and Custom Diffusion are fixed at $3\times10^{-5}$ and the training steps of these methods are fixed to $500$. The learning rates of LoRA and Textual Inversion are fixed at $5\times10^{-5}$ and $1\times10^{-4}$ respectively. The training steps of LoRA and Textual Inversion are fixed at $300$ and $3000$ respectively. We make sure that with such fine-tuning settings, Stable Diffusion can successfully learn the concept from clean datasets. All these fine-tuning methods are based on the \textit{diffusers} library\footnote{\url{https://github.com/huggingface/diffusers}}. The prompts used for fine-tuning are "a photo of a \textit{S*} person" and "a painting in the style of \textit{S*}" for CelebA-HQ and WikiArt datasets respectively. 

\subsection{Settings of Perturbations}
 All implementations of the protection methods are based on their official code and websites\footnote{\url{https://github.com/VinAIResearch/Anti-DreamBooth}}\footnote{\url{https://github.com/CaradryanLiang/ImprovedAdvDM}}\footnote{\url{https://glaze.cs.uchicago.edu/}}. We maintain a fixed perturbation scale of 8/255 for $\ell_\infty$ noise (AdvDM, Anti-DreamBooth, and Improved-AdvDM). We set the optimizing rate of perturbations to $2/255$ and the number of steps to $100$ for AdvDM and the number of iterations to 10 for Anti-DreamBooth to ensure that the perturbations can provide enough protection. Additionally, we apply an adequate amount of Glaze perturbations to images following the recommended settings of its official application.

\subsection{Settings of Natural Transformation}
We apply two simple natural transformations to the protected image, including Gaussian blur and JPEG compression. For Gaussian blur, the kernel size is set to $7\times7$ and $\sigma$ is set to $1.5$. For JPEG compression, we use the implementation of \textit{opencv2} library\footnote{\url{https://github.com/opencv/opencv}} and the compression ratio is set to 40.

\subsection{Settings of Purification}
We follow the official code of DiffPure\footnote{\url{https://github.com/NVlabs/DiffPure}} with the off-the-shelf unconditional diffusion model trained on ImageNet to purify images and maintain most of the parameters but only change the number of purification steps to 50 or 100 (the total step of UDM is 1000). Experiments in Section~\ref{subsec:diffpure} apply 100 steps DiffPure to ensure that the perturbations are successfully removed. For GrIDPure, we fix the number of iterations at 10 and the purification steps in each GDP iteration at 10 and $\gamma$ at 0.1, which are sufficient to remove the protective perturbations.

\section{Additional Results}
\label{sec:results}
In this section, we demonstrate more visulization results of Section~\ref{sec:evaluate} and Section~\ref{sec:defense}.

\subsection{Different Fine-tuning Methods}
\begin{figure*}
  \centering
\includegraphics[width=0.9\linewidth]{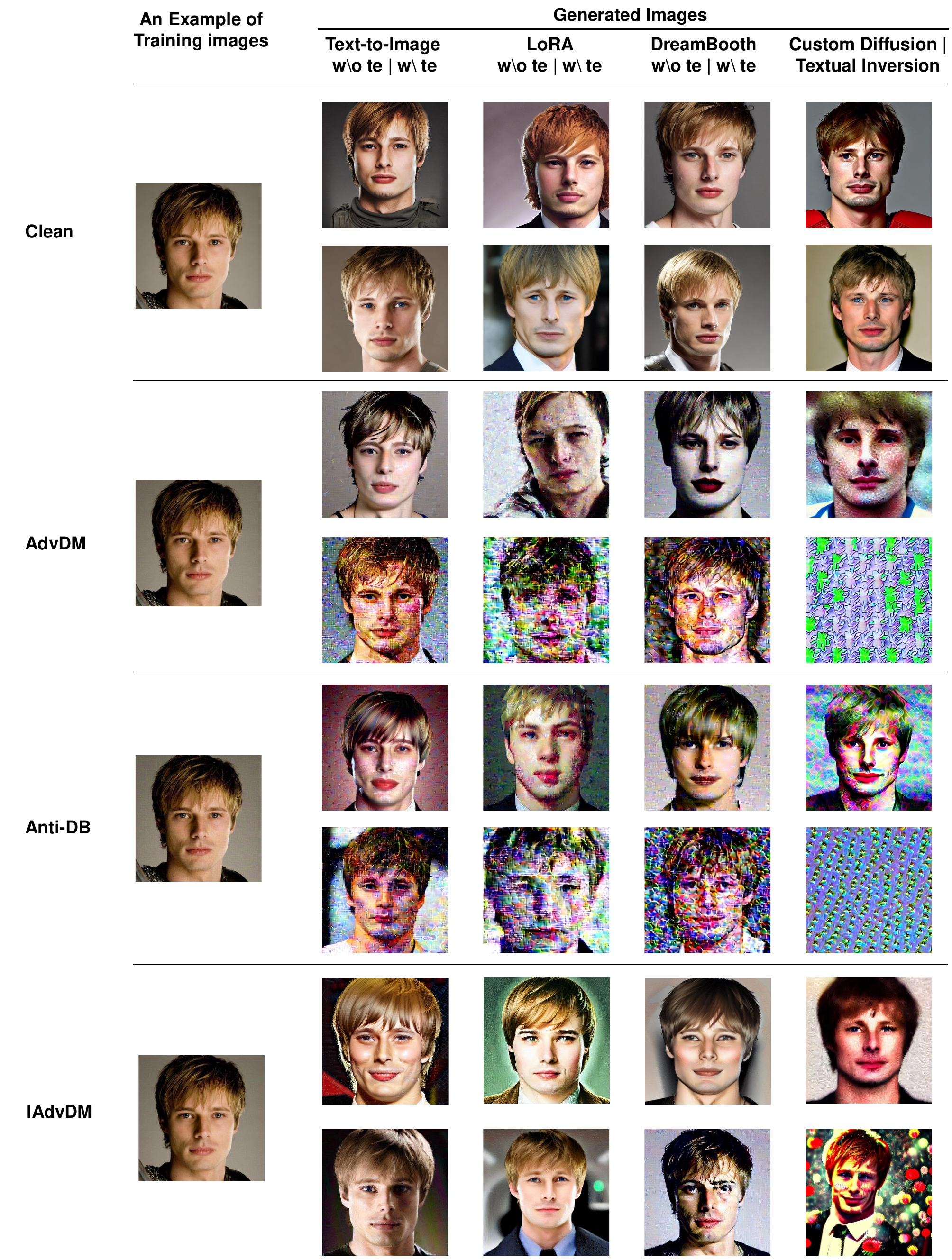}
\caption{Generated images of Stable Diffusion training with different fine-tuning methods and different protective datasets. The prompt of generating is \textit{"a photo of a sks person"}.}
\label{fig:FT_app1}
\end{figure*}

\begin{figure*}
  \centering
\includegraphics[width=0.9\linewidth]{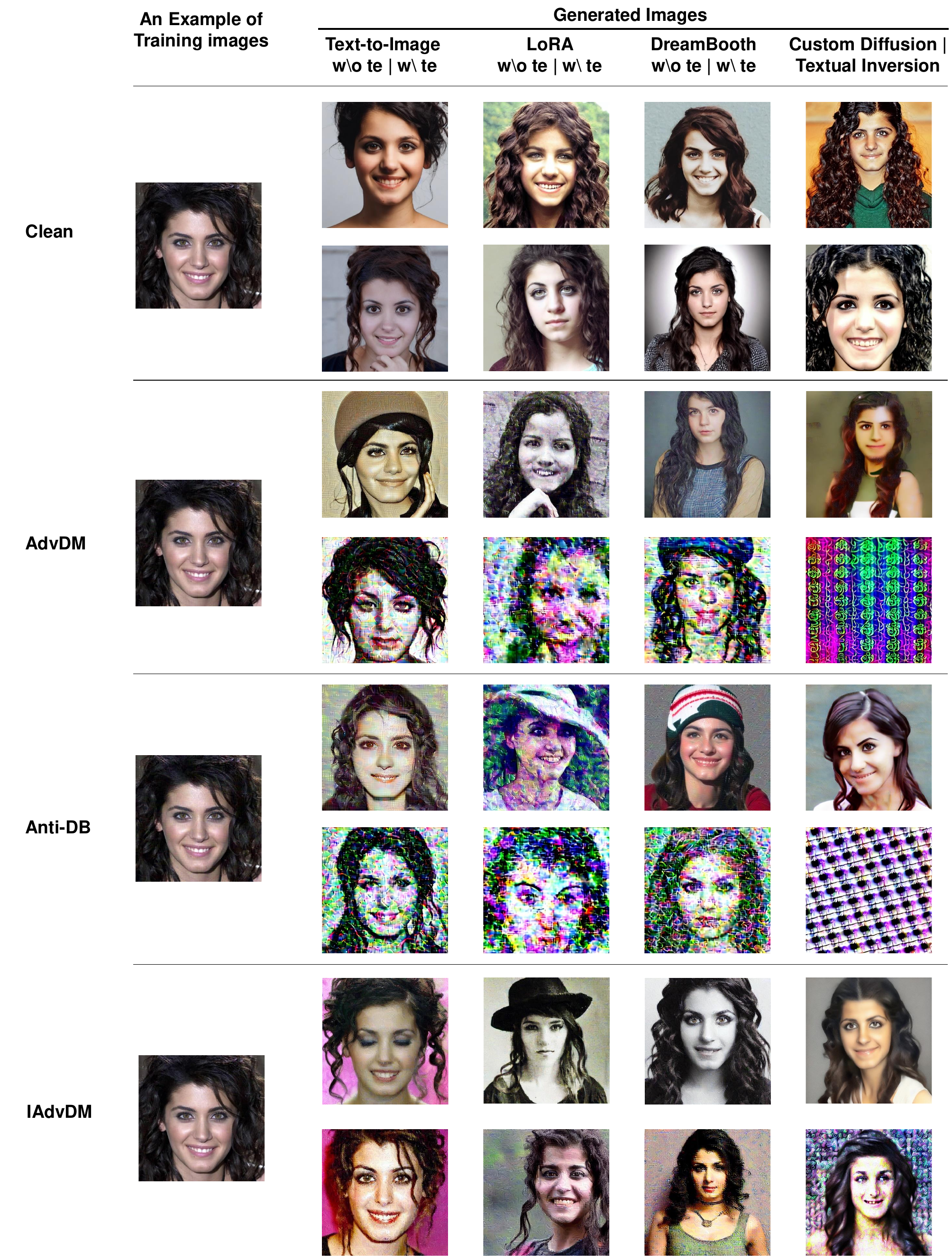}
\caption{Generated images of Stable Diffusion training with different fine-tuning methods and different protective datasets. The prompt of generating is \textit{"a photo of a sks person"}.}
\label{fig:FT_app2}
\end{figure*}

\begin{figure*}
  \centering
\includegraphics[width=0.9\linewidth]{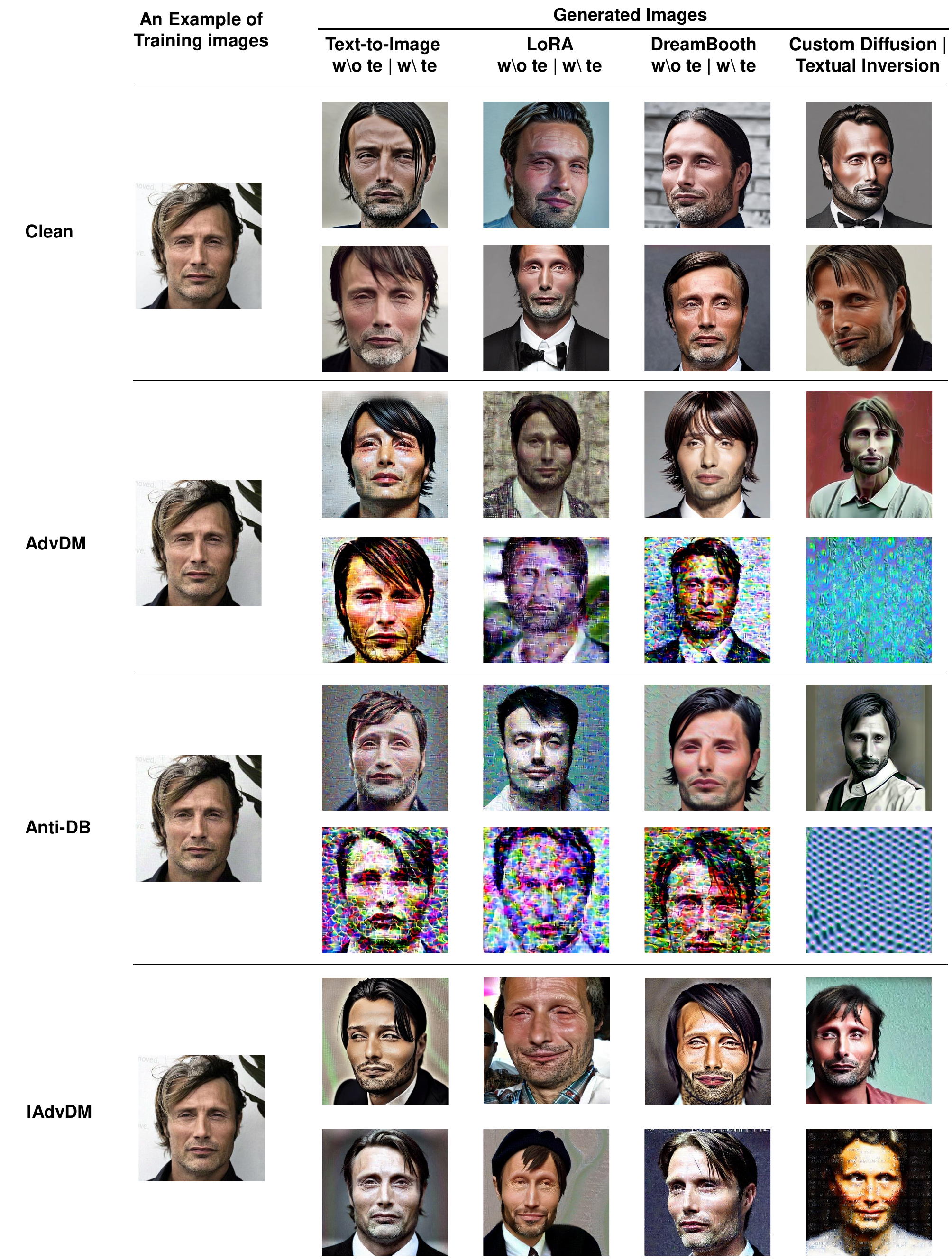}
\caption{Generated images of Stable Diffusion training with different fine-tuning methods and different protective datasets. The prompt of generating is \textit{"a photo of a sks person"}.}
\label{fig:FT_app3}
\end{figure*}

\begin{figure*}
  \centering
\includegraphics[width=0.9\linewidth]{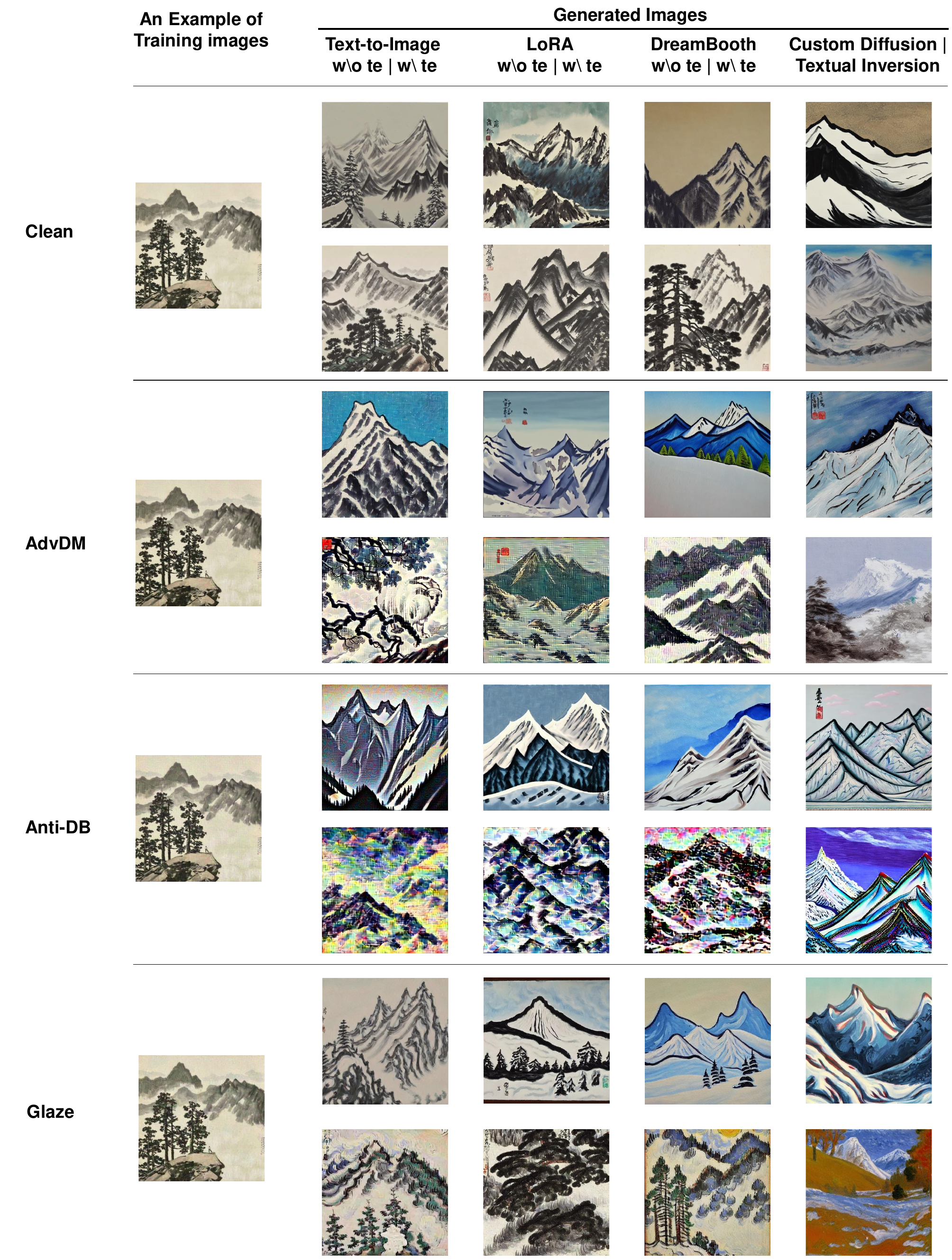}
\caption{Generated images of Stable Diffusion training with different fine-tuning methods and different protective datasets. The prompt of generating is \textit{"a painting of snow mountain in the style of Xu Bei-hong"}.}
\label{fig:FT_app4}
\end{figure*}

\begin{figure*}
  \centering
\includegraphics[width=0.9\linewidth]{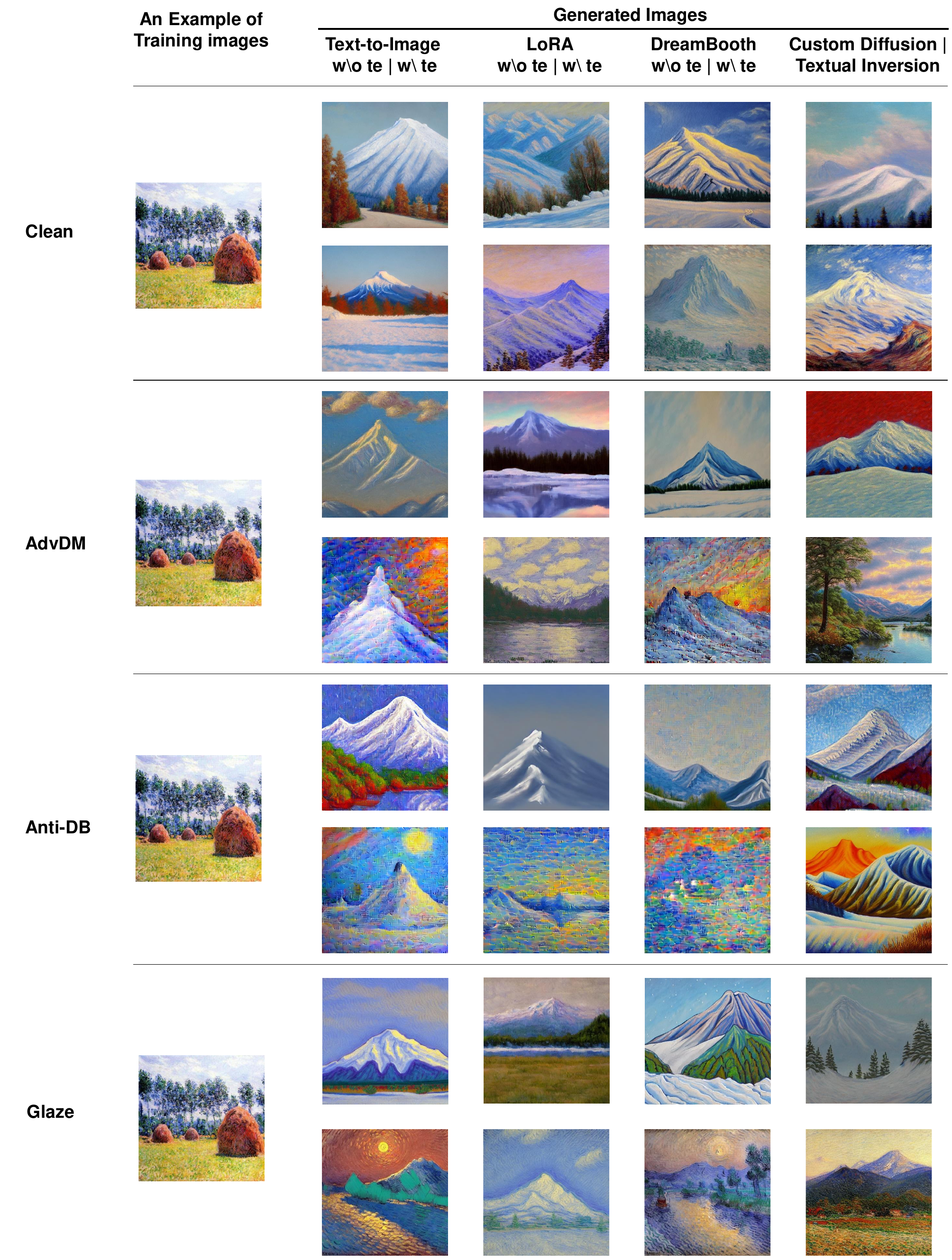}
\caption{Generated images of Stable Diffusion training with different fine-tuning methods and different protective datasets. The prompt of generating is \textit{"a painting of snow mountain in the style of Monet"}.}
\label{fig:FT_app5}
\end{figure*}

Results in Figure~\ref{fig:FT_app1}, Figure~\ref{fig:FT_app2}, Figure~\ref{fig:FT_app3}, Figure~\ref{fig:FT_app4} and Figure~\ref{fig:FT_app5} show more visualization of the effectiveness of different protective perturbations and different fine-tuning methods. The first, third, fifth and seventh lines are the results of without fine-tuning the text encoder and the other lines are the results of fine-tuning the text encoder. These indicate that the performance of protective perturbations is highly related to the chosen fine-tuning methods of image exploiters, especially the methods that train the text encoder.

\subsection{Purification}
\begin{figure*}
  \centering
\includegraphics[width=0.95\linewidth]{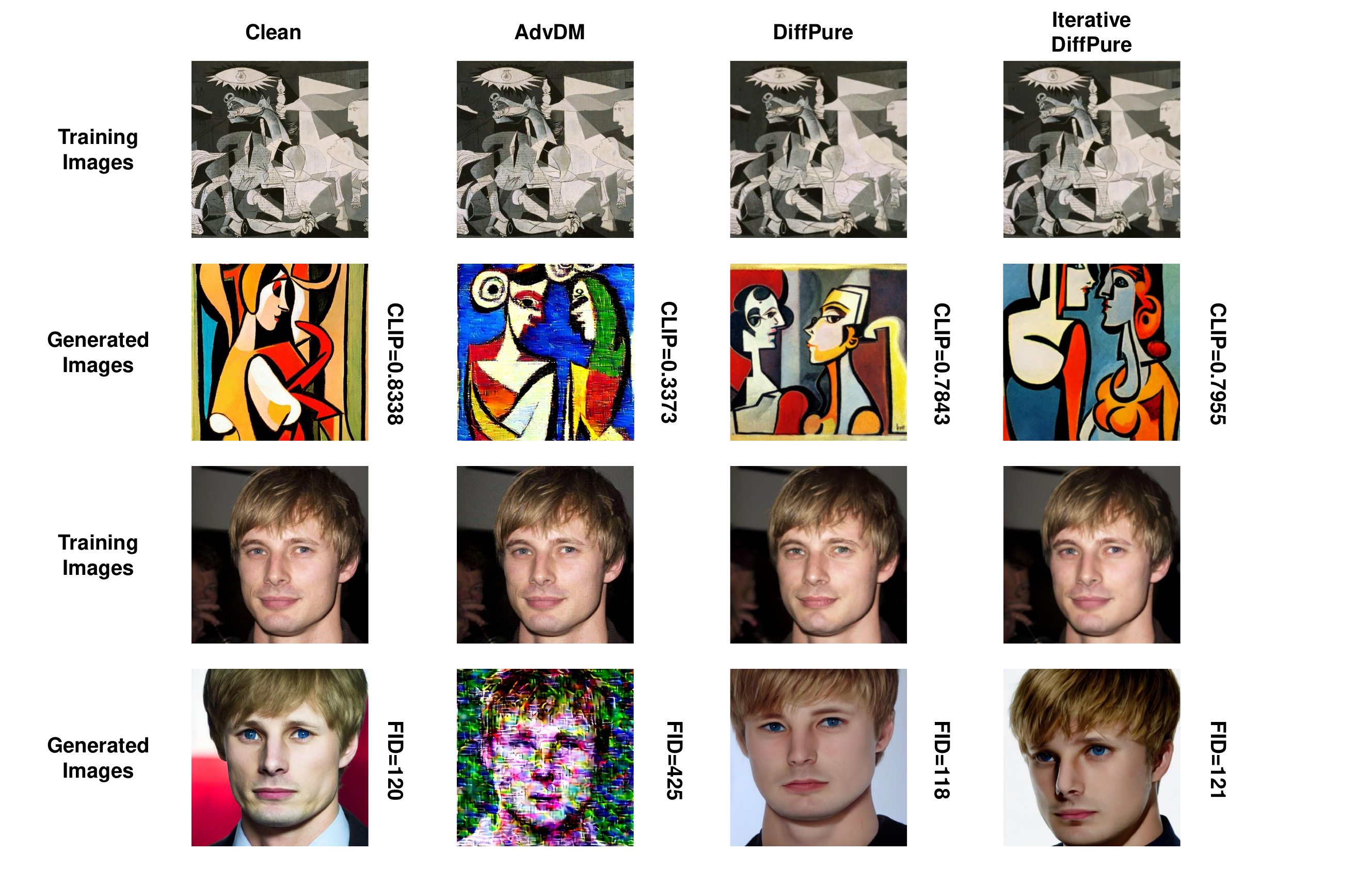}
\caption{Iterative DiffPure with small steps can also successfully bypass the protective perturbations. The 100-step DiffPure is broken down into 10 iterations of 10-step DiffPure.}
\label{fig:IDP}
\end{figure*}

\paragraph{Iterative DiffPure with Small Steps.}
The example in Figure~\ref{fig:IDP} shows that protective perturbation can be removed by iterating a small-step DiffPure multiple times. Considering that a small-step purification changes less structure of the original clean images, we design our GrIDPure based on this insight.

\begin{figure*}
  \centering
\includegraphics[width=0.9\linewidth]{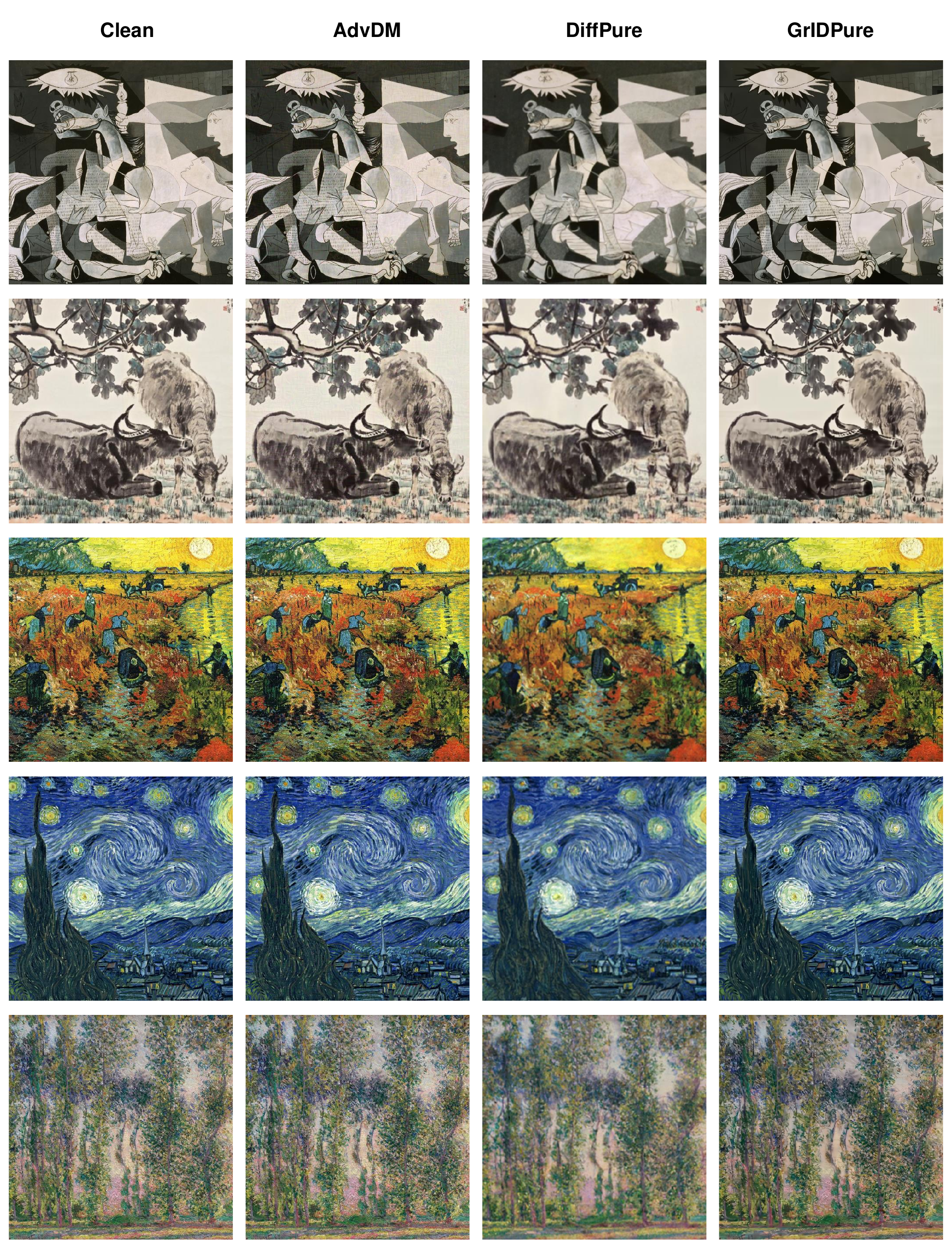}
\caption{Visualization of clean training images, AdvDM-protected images and images purified by DiffPure and GrIDPure. Our GrIDPure can better preserve the quality (resolution and structure) of the original clean image.}
\label{fig:pq1}
\end{figure*}

\begin{figure*}
  \centering
\includegraphics[width=0.9\linewidth]{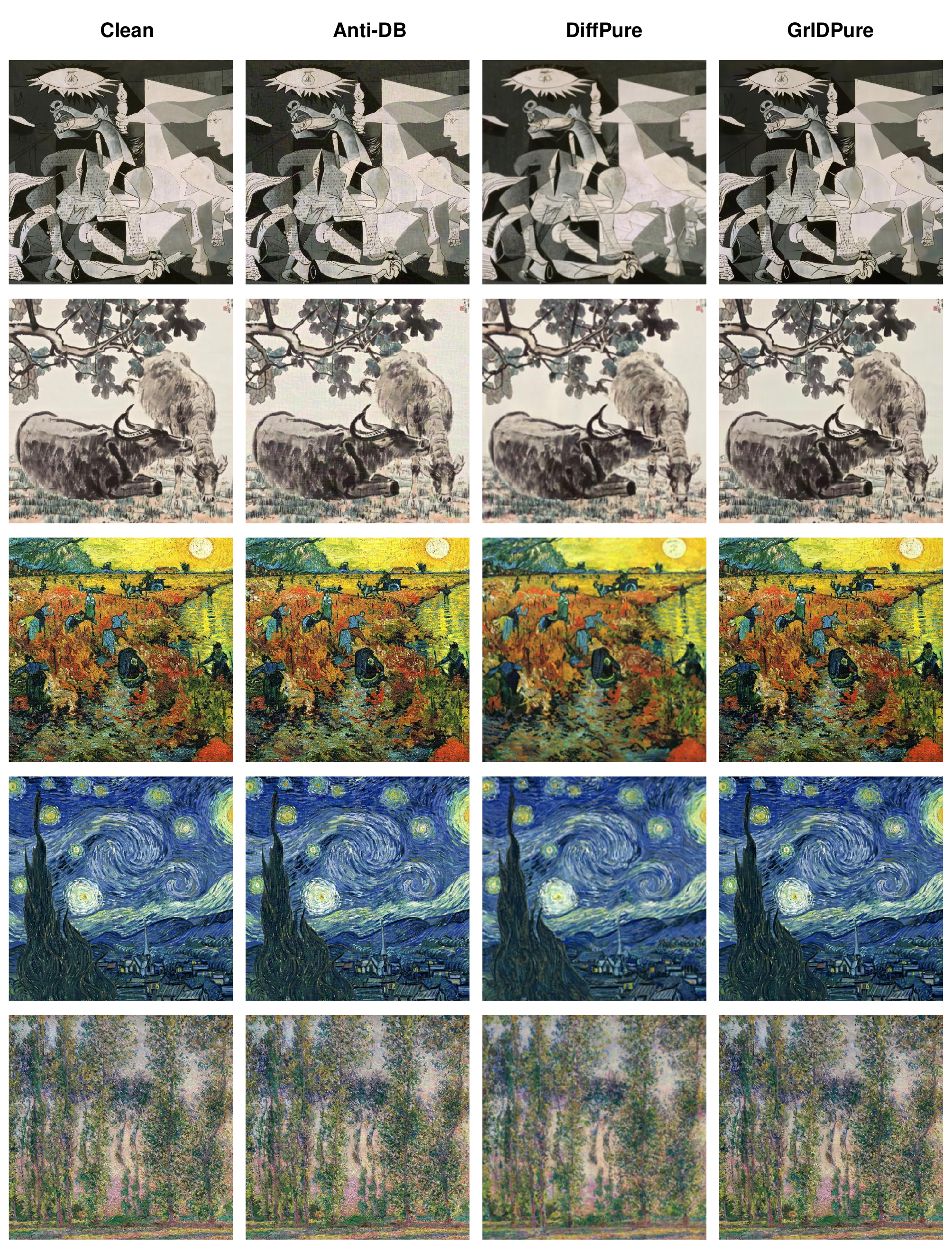}
\caption{Visualization of clean training images, AntiDB-protected images and images purified by DiffPure and GrIDPure. Our GrIDPure can better preserve the quality (resolution and structure) of the original clean image.}
\label{fig:pq2}
\end{figure*}

\begin{figure*}
  \centering
\includegraphics[width=0.9\linewidth]{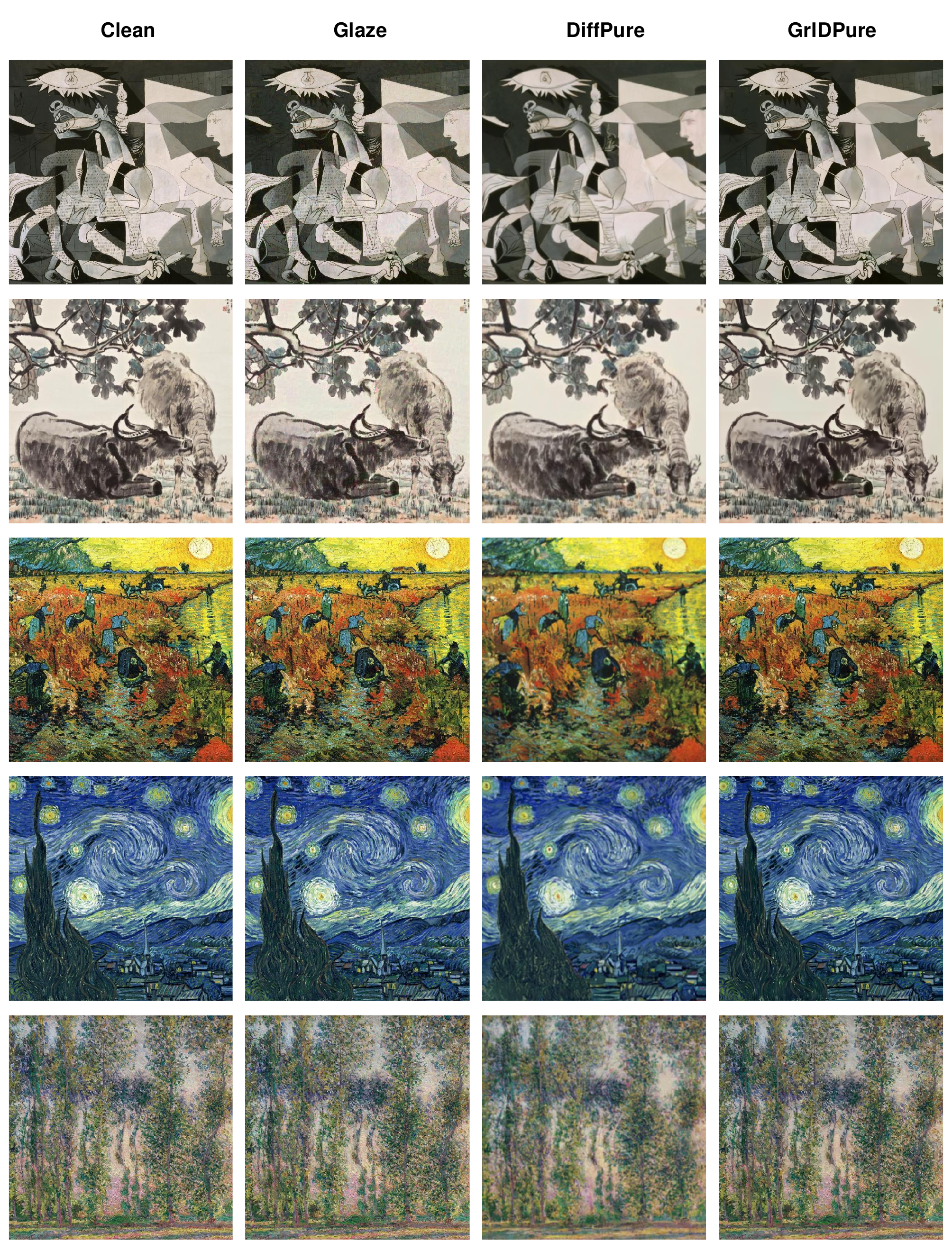}
\caption{Visualization of clean training images, Glaze-protected images and images purified by DiffPure and GrIDPure. Our GrIDPure can better preserve the quality (resolution and structure) of the original clean image.}
\label{fig:pq3}
\end{figure*}

\begin{figure*}
  \centering
\includegraphics[width=0.9\linewidth]{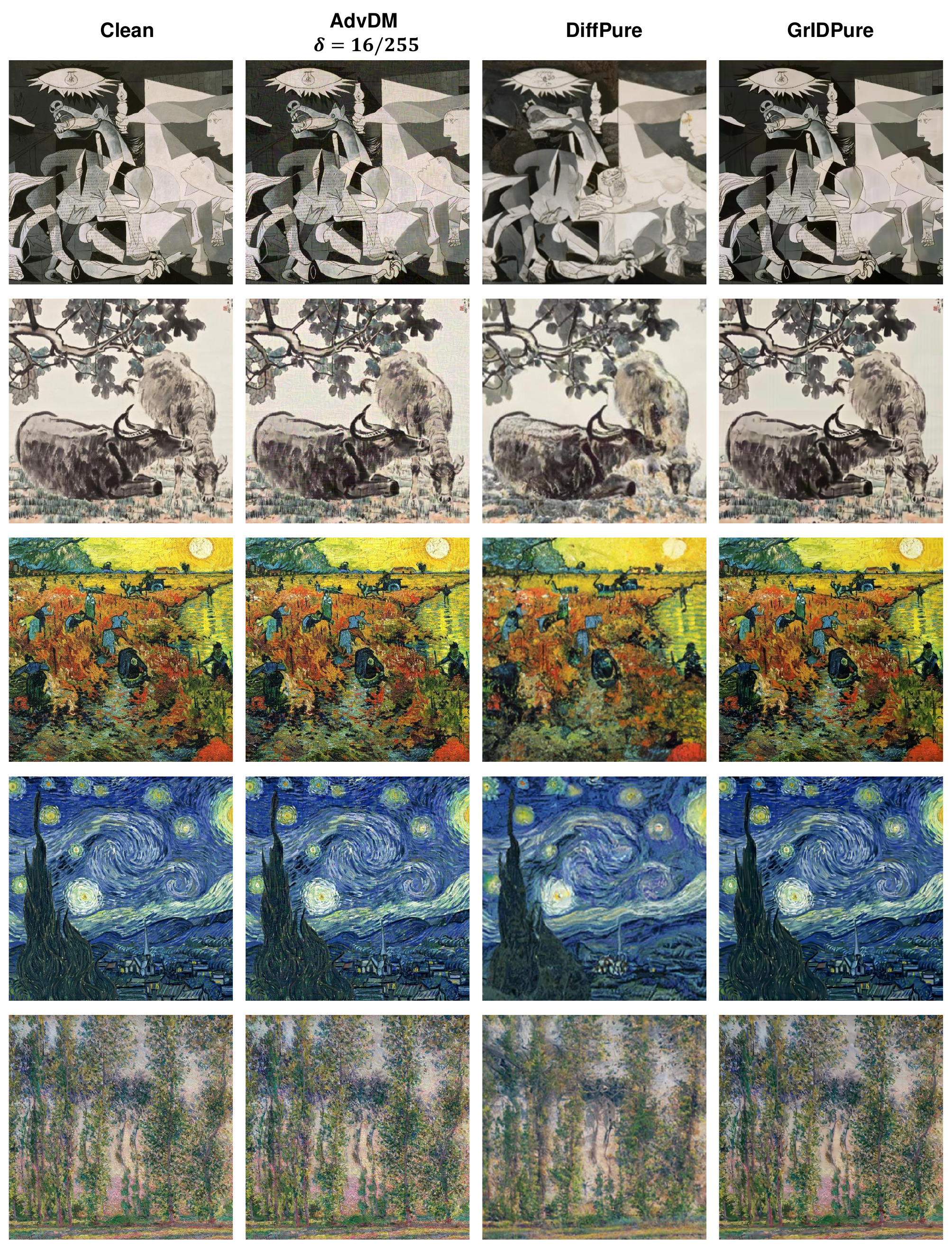}
\caption{Visualization of clean training images, AdvDM-protected ($\delta=16/255$) images and images purified by DiffPure and GrIDPure. Our GrIDPure can better preserve the quality (resolution and structure) of the original clean image.}
\label{fig:pq4}
\end{figure*}

\vspace{-10pt}
\paragraph{Quality of Purification.}
Results in Figure~\ref{fig:pq1}, Figure~\ref{fig:pq2}, Figure~\ref{fig:pq3} and Figure~\ref{fig:pq4} compare the quality of purification between DiffPure and our GrIDPure. We set the purification steps of DiffPure to 100, and the purification steps and the number of iterations of GrIDPure to 10 and 10 respectively in Figure~\ref{fig:pq1}, Figure~\ref{fig:pq2} and Figure~\ref{fig:pq3}. We set the purification steps of DiffPure to 200, and the purification steps and the number of iterations of GrIDPure to 10 and 20 respectively in Figure~\ref{fig:pq4} to ensure the complete purification. 

\begin{figure*}
  \centering
\includegraphics[width=0.9\linewidth]{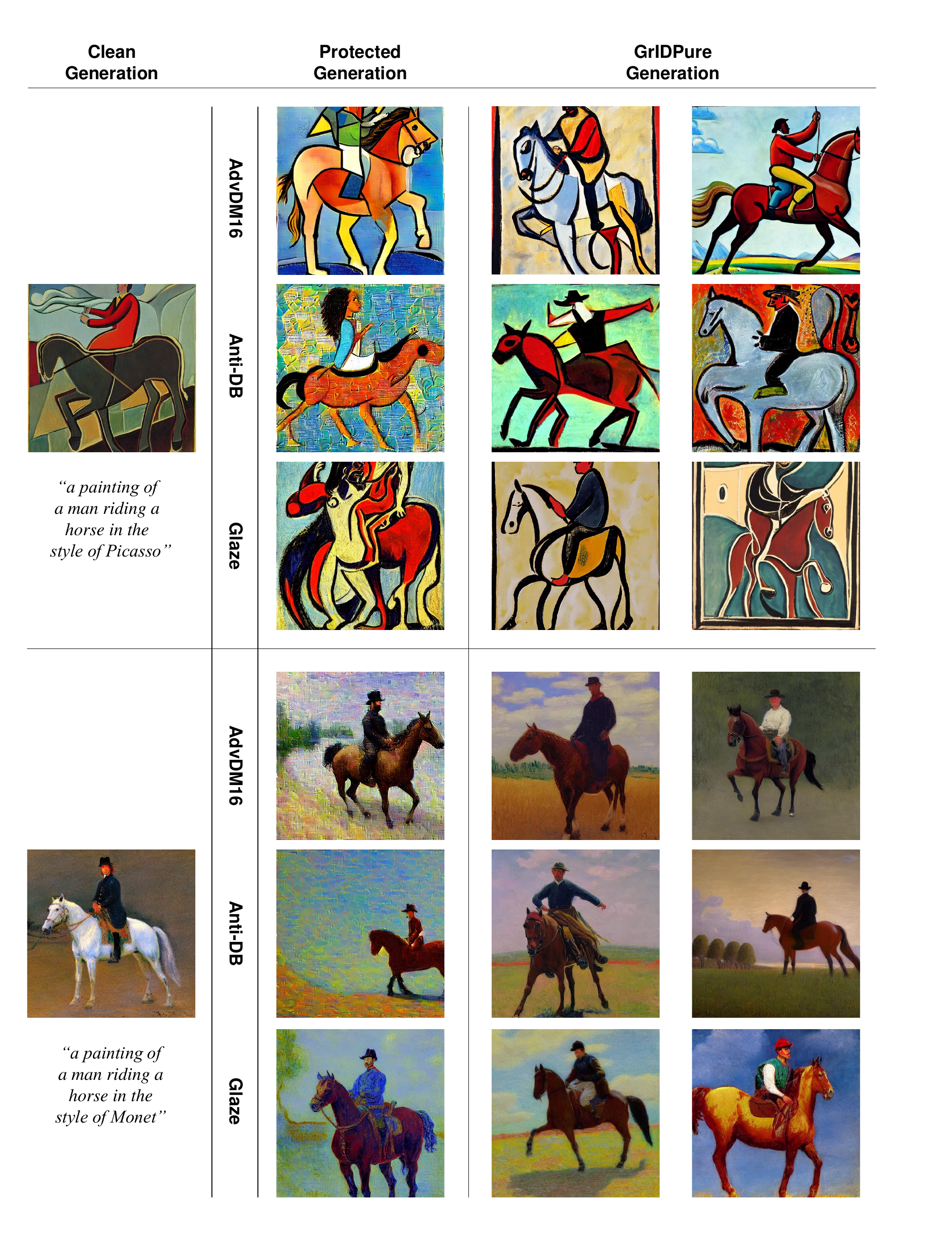}
\caption{Visualization of generated images by Stable Diffusion fine-tuning with images purified by GrIDPure.}
\label{fig:pe1}
\end{figure*}

\begin{figure*}
  \centering
\includegraphics[width=0.9\linewidth]{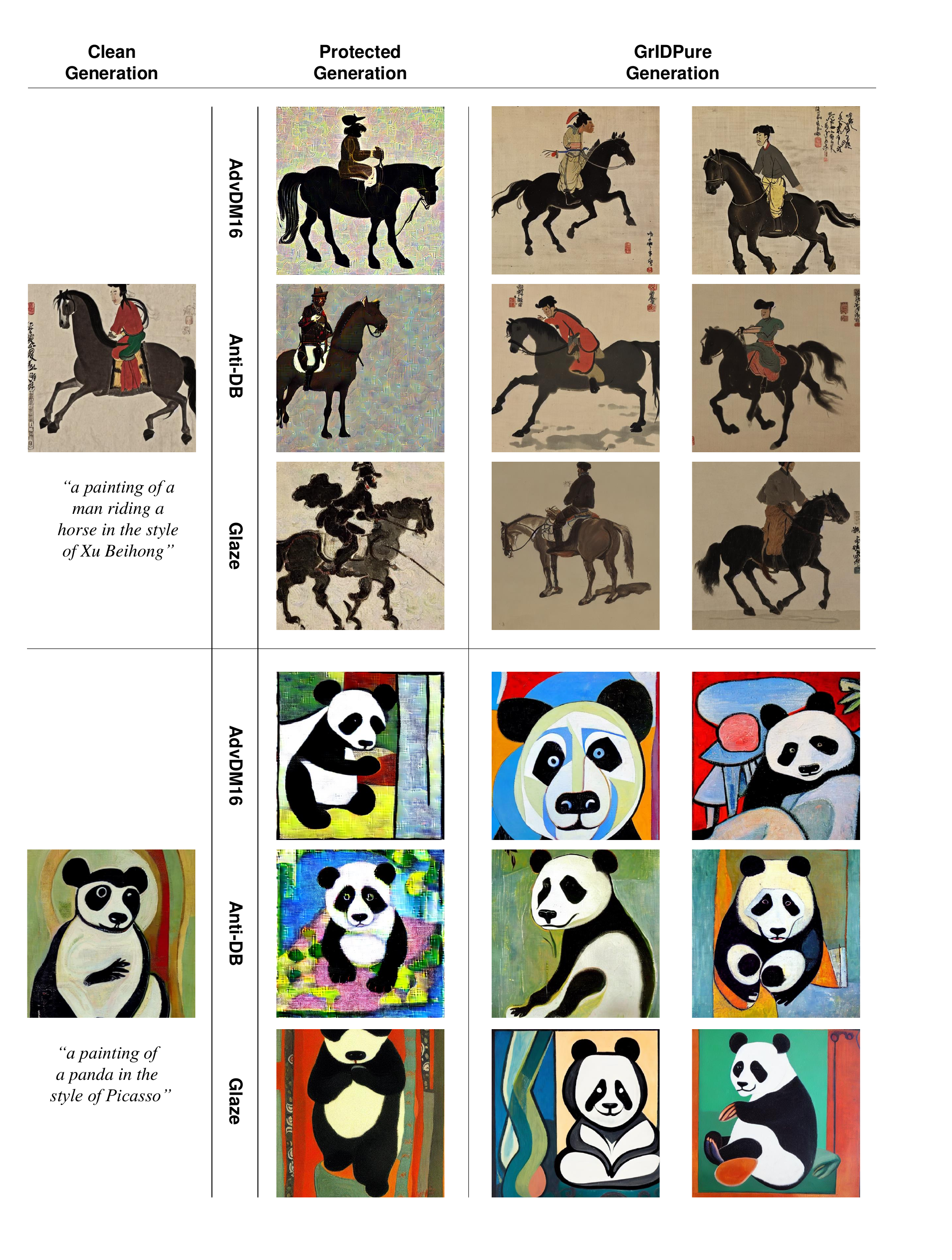}
\caption{Visualization of generated images by Stable Diffusion fine-tuning with images purified by GrIDPure.}
\label{fig:pe2}
\end{figure*}

\begin{figure*}
  \centering
\includegraphics[width=0.9\linewidth]{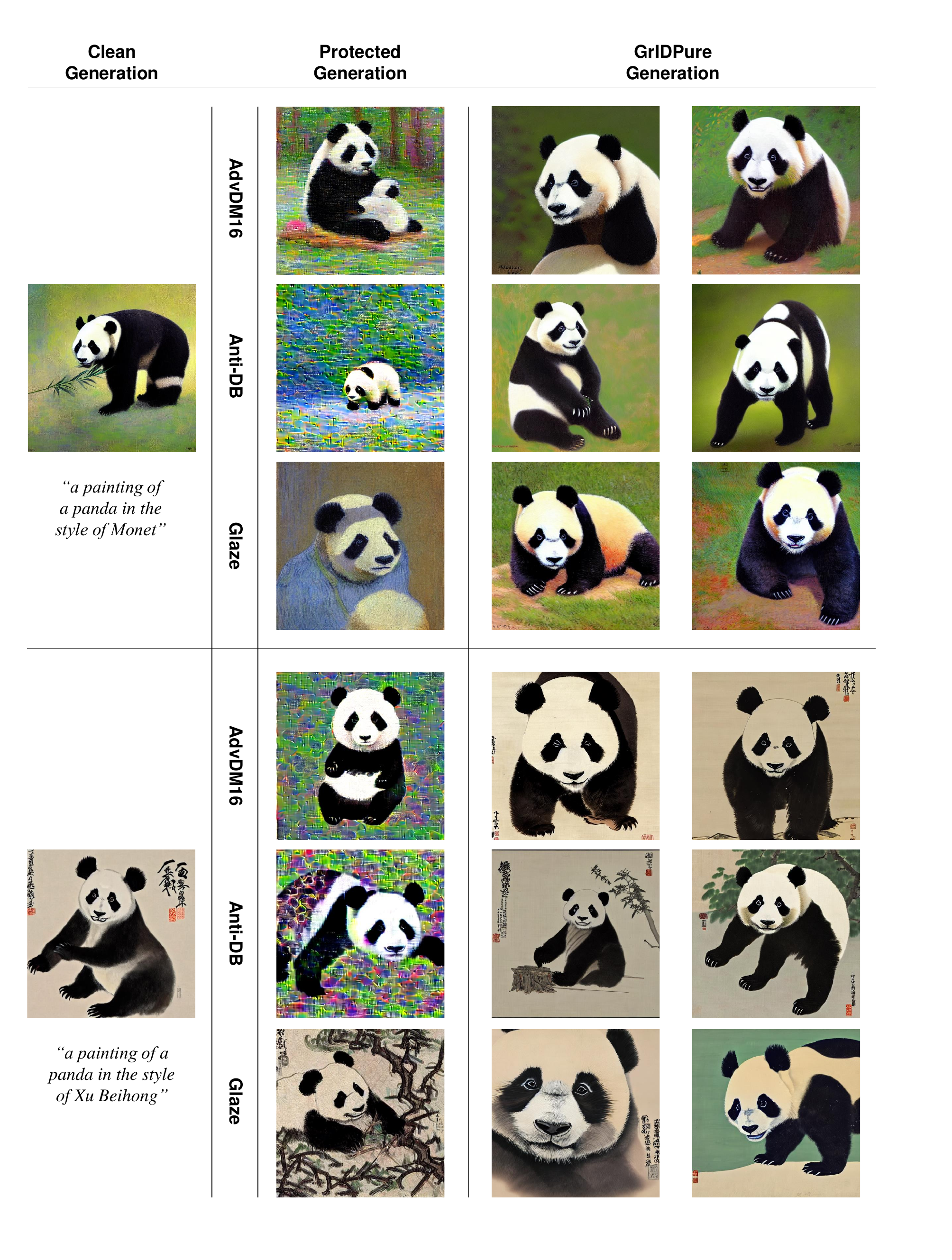}
\caption{Visualization of generated images by Stable Diffusion fine-tuning with images purified by GrIDPure.}
\label{fig:pe3}
\end{figure*}

\begin{figure*}
  \centering
\includegraphics[width=0.9\linewidth]{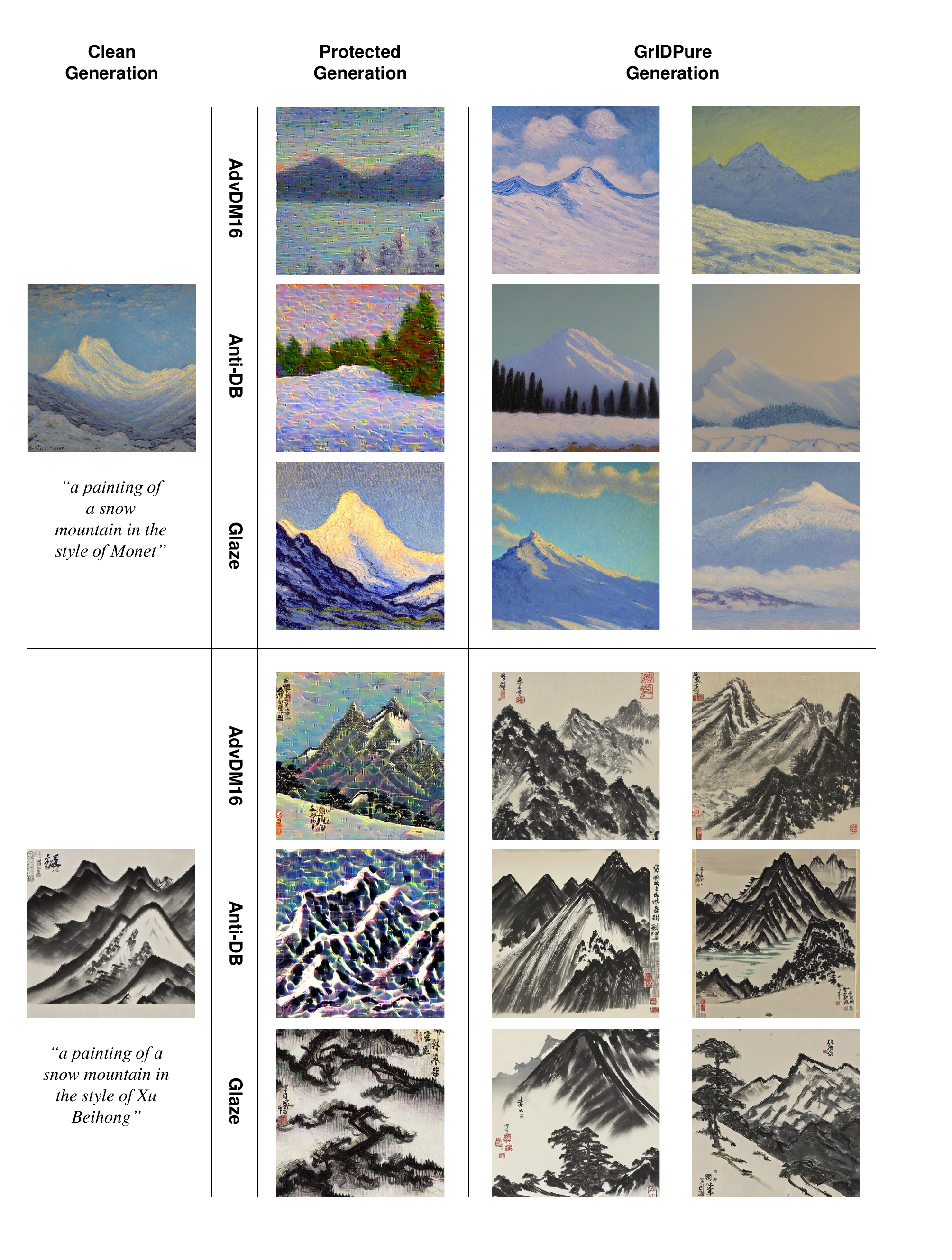}
\caption{Visualization of generated images by Stable Diffusion fine-tuning with images purified by GrIDPure.}
\label{fig:pe4}
\end{figure*}

\begin{figure*}
  \centering
\includegraphics[width=0.9\linewidth]{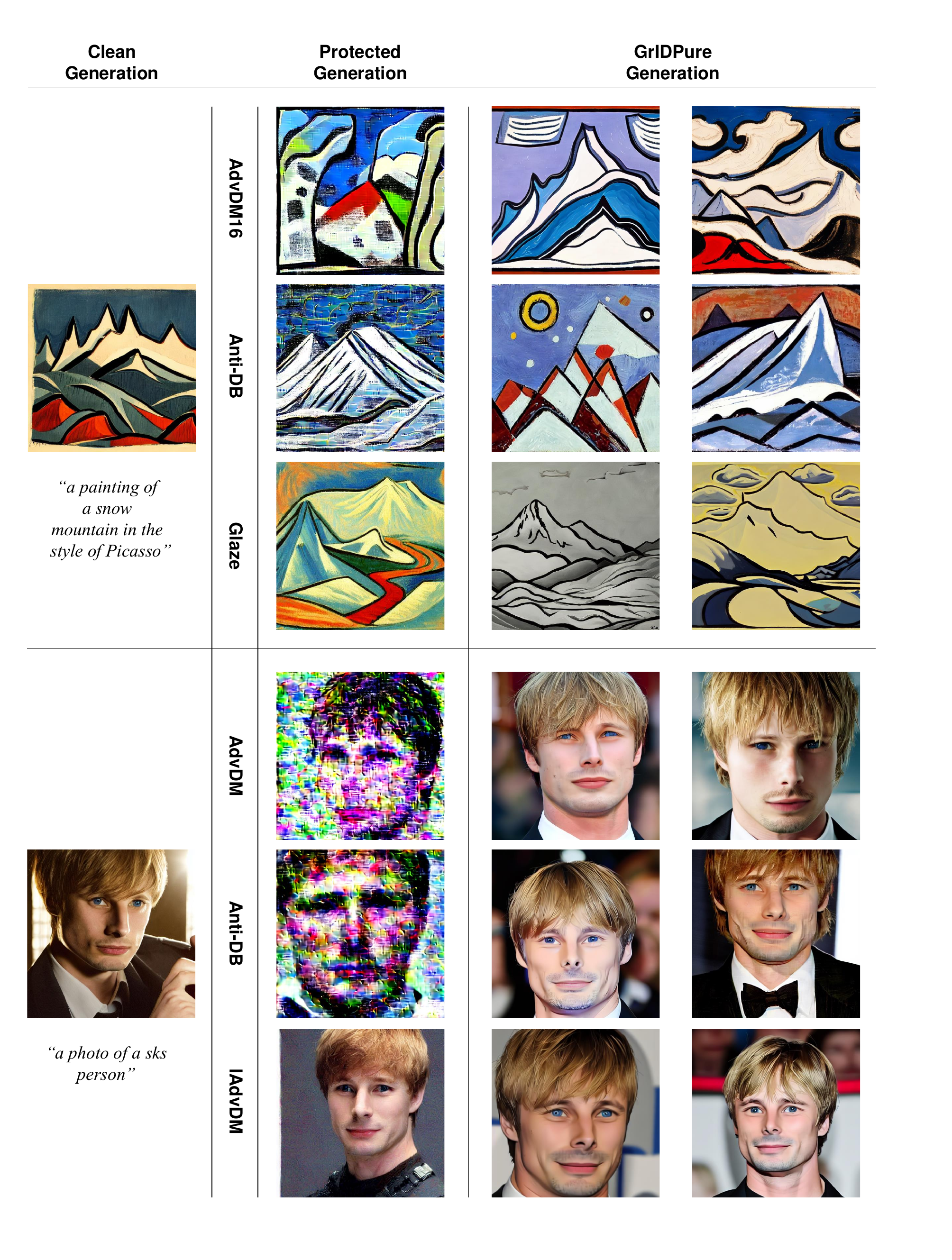}
\caption{Visualization of generated images by Stable Diffusion fine-tuning with images purified by GrIDPure.}
\label{fig:pe5}
\end{figure*}

\begin{figure*}
  \centering
\includegraphics[width=0.9\linewidth]{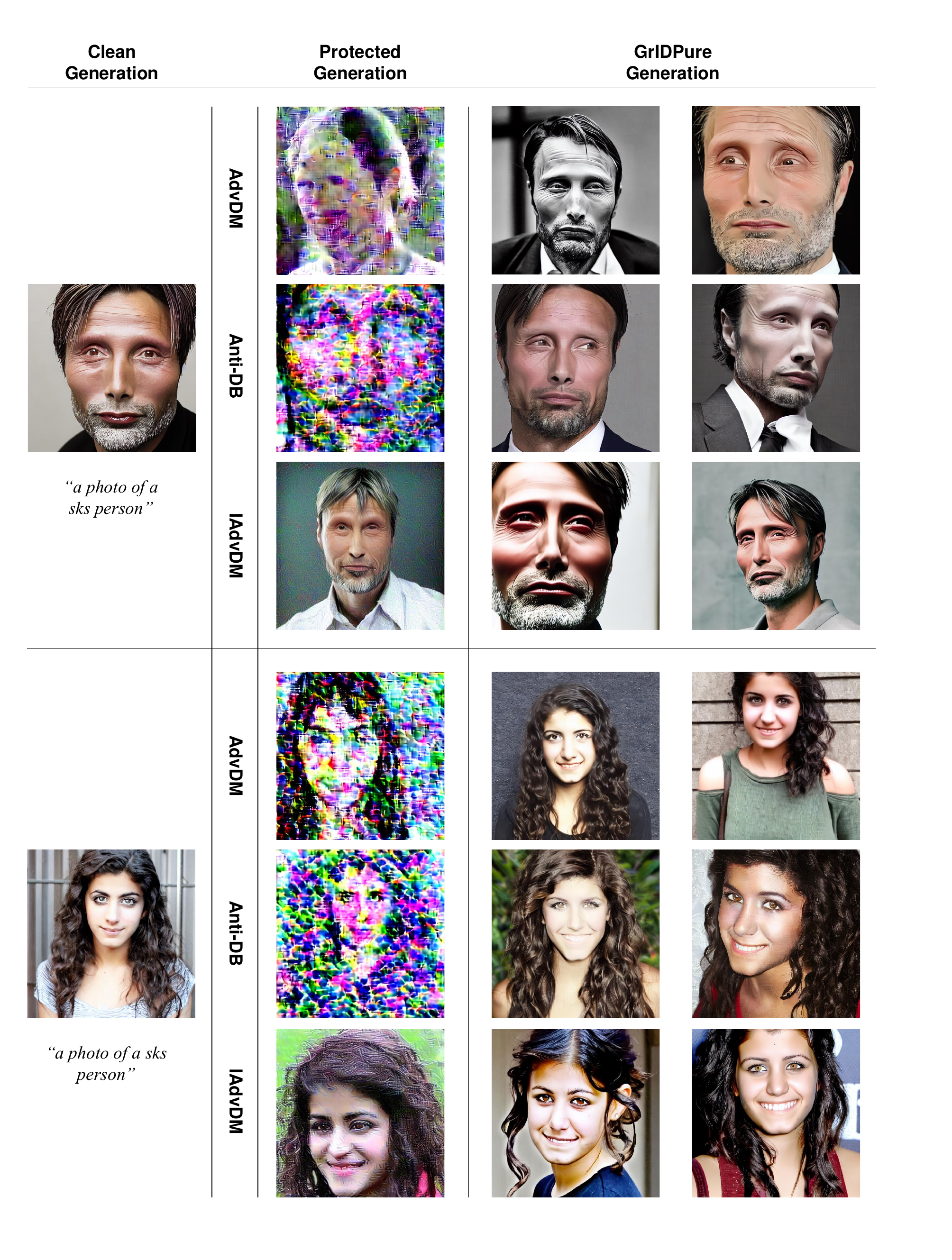}
\caption{Visualization of generated images by Stable Diffusion fine-tuning with images purified by GrIDPure.}
\label{fig:pe6}
\end{figure*}

 \begin{table*}[]
\centering
\footnotesize
\setlength{\tabcolsep}{1pt}
\begin{tabular}{c|c|c|ccc|ccc}
\toprule
\begin{tabular}[c]{@{}c@{}}Training\\ Method\end{tabular} & Metrics & Clean & AdvDM & \begin{tabular}[c]{@{}c@{}}AdvDM\\ +DiffPure\end{tabular} & \begin{tabular}[c]{@{}c@{}}AdvDM\\ +GrIDPure\end{tabular} & AntiDB & \begin{tabular}[c]{@{}c@{}}AntiDB\\ +DiffPure\end{tabular} & \begin{tabular}[c]{@{}c@{}}AntiDB\\ +GrIDPure\end{tabular} \\ \hline
\multirow{2}{*}{LORA w\textbackslash{}te} & FDFR& 0.04  & 1.0& 0.06 & \textbf{0.04} & 0.96& 0.04  & \textbf{0.0}\\ \cline{2-9} 
 & ISM & 0.59  & 0.0& 0.55 & \textbf{0.56} & 0.03& 0.56  & \textbf{0.56}  \\ \hline
\multirow{2}{*}{LORA w\textbackslash{}o te}& FDFR& 0.07  & 0.26  & 0.16 & \textbf{0.10} & 0.18& 0.16  & \textbf{0.06}  \\ \cline{2-9} 
 & ISM & 0.58  & 0.49  & 0.53 & \textbf{0.60} & 0.52& 0.57  & \textbf{0.61}  \\ \bottomrule
\end{tabular}
\caption{More metrics on face-generation tasks.}
\label{tab:fdfr_ism}
\end{table*}
\vspace{-10pt}
\paragraph{Effectiveness of Purification.}
Results in Figure~\ref{fig:pe1}, Figure~\ref{fig:pe2}, Figure~\ref{fig:pe3}, Figure~\ref{fig:pe4}, Figure~\ref{fig:pe5} and Figure~\ref{fig:pe6} demonstrate that our GrIDPure can successfully bypass all SOTA protective perturbations which remove the perturbation on the training images and recover these images into learnable images. The scales of perturbations are set to $8/255$ for AdvDM, Anti-DreamBooth and ImprovedAdvDM, and $16/255$ for AdvDM16. For Glaze, we apply the strongest settings provided by its official application.

\begin{table}[]
\centering
\footnotesize
\setlength{\tabcolsep}{3pt}
\begin{tabular}{c|c|ccccc}
\toprule
Dataset  & Metrcs & Clean  & AdvDM  & Ada.& \begin{tabular}[c]{@{}c@{}}AdvDM\\ +GrIDPure\end{tabular} & \begin{tabular}[c]{@{}c@{}}Ada.\\ +GrIDPure\end{tabular} \\ \hline
\multirow{2}{*}{CelebA}  & FID& 119.8  & 424.7  & 253.1  & 121.4  & 114.8 \\ \cline{2-7} 
  & CLIP& 0.7378 & 0.2316 & 0.5473 & 0.8526 & 0.7406\\ \hline
\multirow{2}{*}{WikiArt} & FID& 201.9  & 251.1  & 240.8  & 203.4  & 206.6 \\ \cline{2-7} 
  & CLIP& 0.8338 & 0.3373 & 0.5124 & 0.8758 & 0.8415\\ \bottomrule
\end{tabular}
\caption{Adaptive attack against GrIDPure.}
\label{tab:ada_gridpure}
\end{table}

\begin{figure}[]
  \centering
\includegraphics[width=0.98\linewidth]{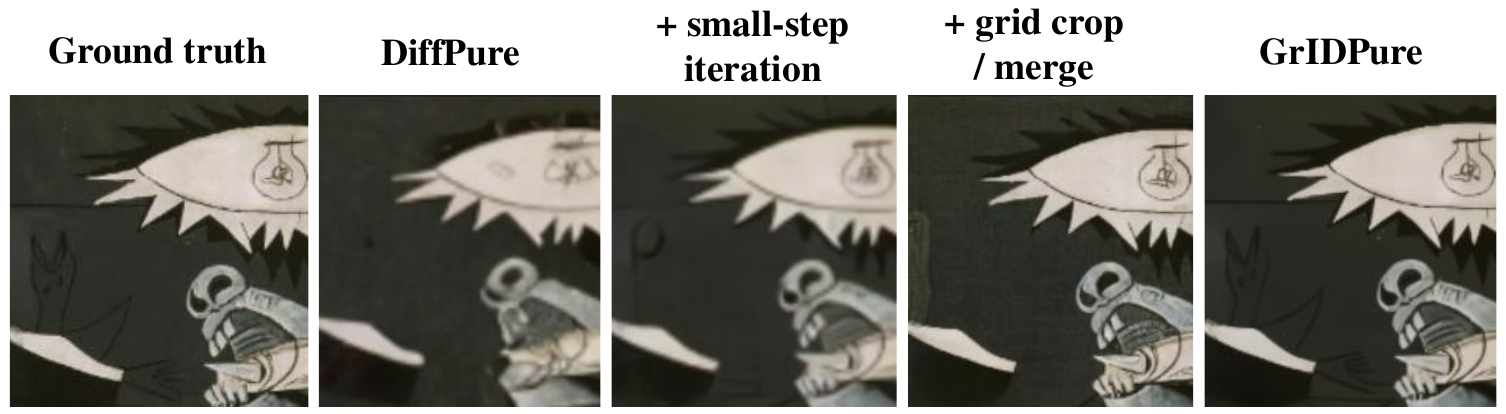}
\caption{Visualization examples of DiffPure, small-step iteration, grid crop/merge and GrIDPure.}
\label{fig:ablation_iter_grid}
\end{figure}

\subsection{Ablation Study} 
\paragraph{Adaptive Attack against GrIDPure}
Considering the theoretical consistency between GrIDPure and DiffPure and the significant GPU memory requirements, it's almost infeasible to conduct Adaptive Attacks on the GrIDPure framework. We make our best effort to design a white-box adaptive attack specifically targeting the core component of GrIDPure, \textit{Small-step Iteration}. Results in Table~\ref{tab:ada_gridpure} indicate that such adaptive attacks do not diminish the effectiveness of GrIDPure.

\begin{figure}[]
  \centering
\includegraphics[width=0.9\linewidth]{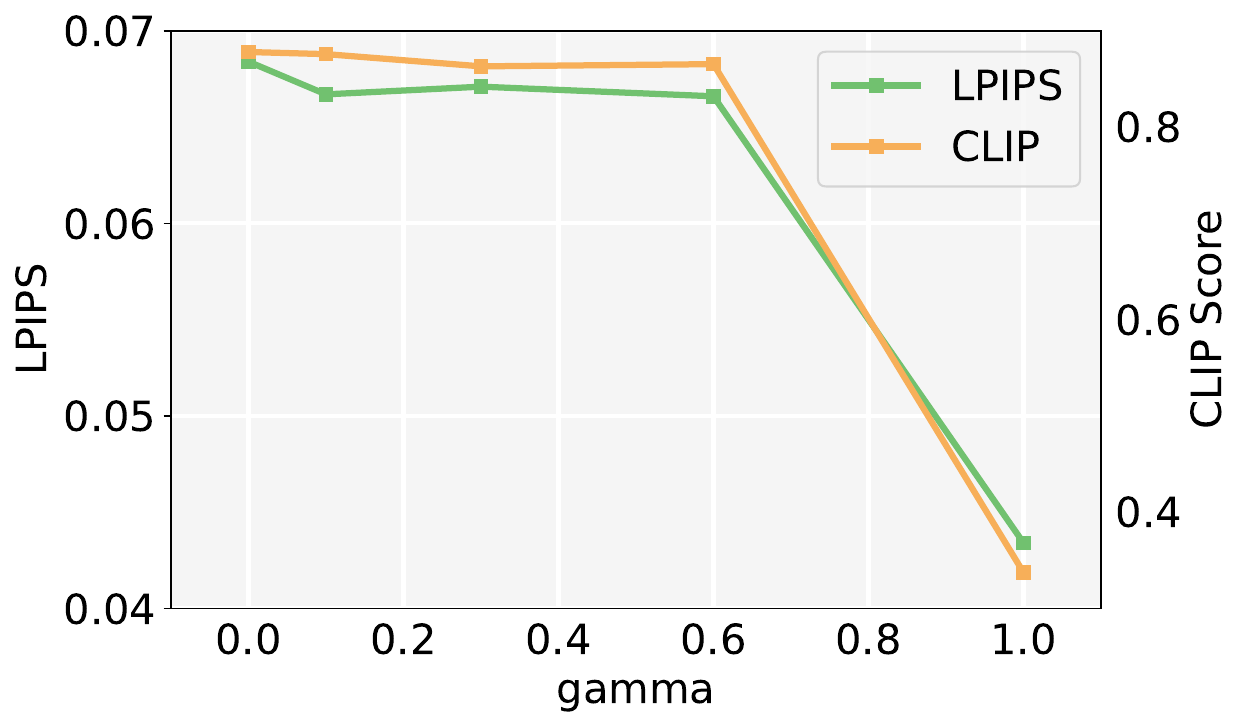}
\caption{Ablation study on blending parameter $\gamma$.}
\label{fig:ablation_gamma}
\end{figure}

\begin{table}[]
\centering
\footnotesize
\begin{tabular}{l|ccc|cc}
\toprule
\multicolumn{1}{c|}{Assessment} & \multicolumn{3}{c|}{Quality} & \multicolumn{2}{c}{Effectiveness} \\ \hline
\multicolumn{1}{c|}{Metrics} & PSNR & SSIM& LPIPS& FID & CLIP\\ \hline
\textit{AdvDM (No Pure.)} & \textit{37.46} & \textit{0.9496} & \textit{0.0434} & \textit{251.1} & \textit{0.3373} \\ \hline
DiffPure & 22.24& 0.6378& 0.4425  & 214.2  & 0.7843 \\ \hline
+ small-step iter.& 23.42& 0.7175& 0.3904  & 211.2  & 0.7955 \\ \hline
+ grid crop/ merge& 27.14& 0.8052& 0.0757  & 214.7  & 0.7577 \\ \hline
GrIDPure & \textbf{30.60}& \textbf{0.9199}& \textbf{0.0672}& \textbf{203.4}& \textbf{0.8758} \\ \bottomrule
\end{tabular}
\caption{Ablation study on small-step iteration and grid crop/merge.}
\label{tab:ablation_iter_grid}
\end{table}

\paragraph{Ablation Studies on GrIDPure}
We do ablation studies for mechanisms (shown in Figure~\ref{fig:ablation_iter_grid} and Table~\ref{tab:ablation_iter_grid}) and blending parameter $\gamma$ (shown in Figure~\ref{fig:ablation_gamma}) in GrIDPure. Small-step Iteration aids in better preserving the details of the images, while grid crop/merge helps in retaining the resolution of the images. As shown in Figure~\ref{fig:ablation_gamma}, by appropriately blending images from different iterations, we can mitigate the loss of details during the SDEdit processes, striking a balance between preserving image details (smaller LPIPS~\cite{Zhang_2018_CVPR}) and removing protective perturbation (higher CLIP-Score).

\subsection{Additional Metrics}
To further demonstrate the influence of protective perturbations on the face-generation task, we refer to Anti-DreamBooth~\cite{van2023anti} to calculate FDFR and ISM for the face generation in Table~\ref{tab:fdfr_ism}, where the lower FDFR and higher ISM represent better generative quality. The results from these metrics align with the conclusions that fine-tuning the text encoder greatly enhances the protection efficacy diffusion-based purification can successfully remove these protections.

\section{Broader Impact}
This paper evaluates methods that use protective perturbations to prevent generative models from exploiting personal data, thereby addressing concerns such as privacy breaches and copyright infringement. Additionally, the paper proposes approaches to bypass these protections, potentially exposing protected data to risks and providing opportunities for unauthorized exploiters to bypass existing protective measures. Despite these challenges, we believe that assessing the effectiveness of such protections is crucial. In the long run, our work holds positive implications for safeguarding personal privacy and copyright in images.

\section*{Acknowledgements}
This work is partially supported by the NSF of China(under Grants U22A2028, 61925208, 62222214, 62102399, 62102398, U20A20227, 62372436, 62302478, 62302482, 62302483, 62302480), CAS Project for Young Scientists in Basic Research(YSBR-029), Youth Innovation Promotion Association CAS and Xplore Prize.


\end{document}